\documentclass[lettersize,journal]{IEEEtran}
\usepackage{amsmath,amsfonts}
\usepackage{algorithmic}
\usepackage{algorithm}
\usepackage{array}
\usepackage[caption=false,font=normalsize,labelfont=sf,textfont=sf]{subfig}
\usepackage{textcomp}
\usepackage{stfloats}
\usepackage{url}
\usepackage{verbatim}
\usepackage{graphicx}
\usepackage{cite}

\usepackage[utf8]{inputenc} 
\usepackage[T1]{fontenc}    
\usepackage{hyperref}       
\usepackage{booktabs}       
\usepackage{microtype}      
\usepackage{xcolor}         
\usepackage{graphics}
\usepackage[capitalize]{cleveref}
\usepackage{caption}
\usepackage{subcaption}
\usepackage{multirow}
\usepackage[inline]{enumitem}

\newcommand{\algoname}{{\tt nf2vec}}
\newcommand{\inrtovec}{{\tt nf2vec}}
\newcommand{\nerfalgoname}{{\tt nf2vec}}

\newcommand{\nf}{\emph{NF}}
\newcommand{\udf}{\emph{UDF}}
\newcommand{\sdf}{\emph{SDF}}
\newcommand{\of}{\emph{OF}}
\newcommand{\nerf}{\emph{RF}}
\newcommand{\pc}{point clouds}

\newcommand{\ie}{\textit{i}.\textit{e}.,}
\newcommand{\eg}{\textit{e}.\textit{g}.,}
\newcommand{\wrt}{\textit{w}.\textit{r}.\textit{t}.}

\begin{document}

\title{Deep Learning on Object-centric 3D Neural Fields}

\author{
Pierluigi Zama Ramirez$^*$
\thanks{$^*$Joint first authorship.},
Luca De Luigi$^*$,
Daniele Sirocchi$^*$\\
Adriano Cardace, Riccardo Spezialetti, Francesco Ballerini, Samuele Salti, Luigi Di Stefano \\
University of Bologna\\
\texttt{\small\{pierluigi.zama,
luca.deluigi4, adriano.cardace2, riccardo.spezialetti\}@unibo.it}
}

\markboth{Journal of \LaTeX\ Class Files,~Vol.~14, No.~8, August~2021}%
{Shell \MakeLowercase{\textit{et al.}}: A Sample Article Using IEEEtran.cls for IEEE Journals}


\maketitle

\begin{abstract}
In recent years, Neural Fields (\nf{}s) have emerged as an effective tool for encoding diverse continuous signals such as images, videos, audio, and 3D shapes. When applied to 3D data, \nf{}s offer a solution to the fragmentation and limitations associated with prevalent discrete representations.
However, given that \nf{}s are essentially neural networks, it remains unclear whether and how they can be seamlessly integrated into deep learning pipelines for solving downstream tasks. This paper addresses this research problem and introduces \algoname{}, a framework capable of generating a compact latent representation for an input \nf{} in a single inference pass.
We demonstrate that \algoname{} effectively embeds 3D objects represented by the input \nf{}s and showcase how the resulting embeddings can be employed in deep learning pipelines to successfully address various tasks, all while processing exclusively \nf{}s. 
We test this framework on several \nf{}s used to represent 3D surfaces, such as unsigned/signed distance and occupancy fields. Moreover, we demonstrate the effectiveness of our approach with more complex \nf{}s that encompass both geometry and appearance of 3D objects such as neural radiance fields.
\end{abstract}


\begin{figure*}
    \centering
    \includegraphics[width=0.95\textwidth]{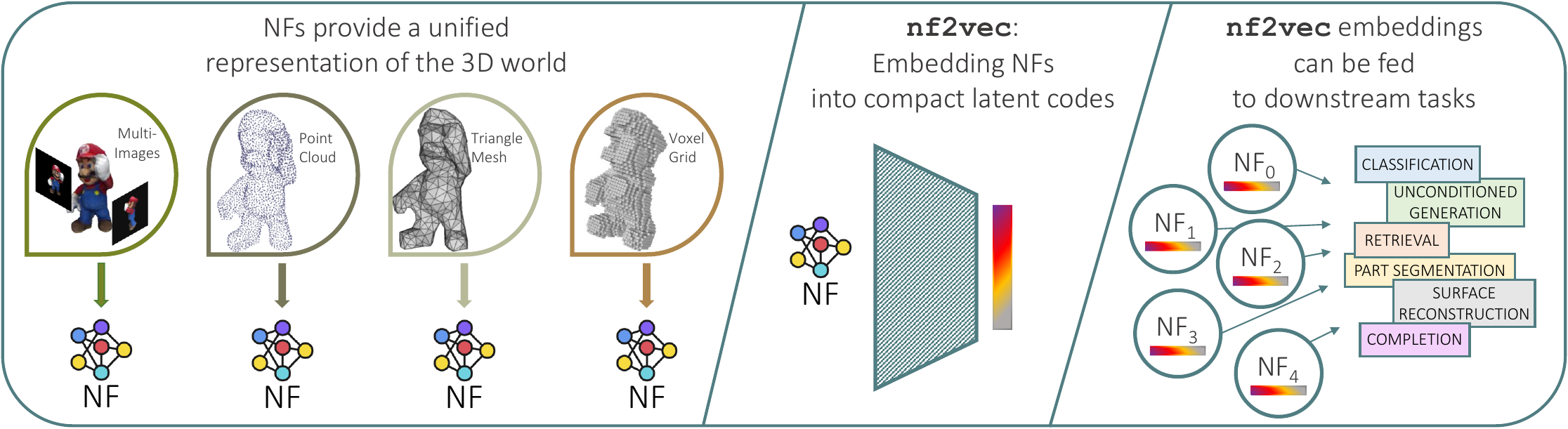}
    \caption{\textbf{Overview of our framework.} \textbf{Left}: \nf{}s hold the potential to provide a unified representation of the 3D world. \textbf{Center}: Our framework, dubbed \algoname{}, produces a compact representation for an input \nf{} by looking only at its weights. \textbf{Right}: \algoname{} embeddings can be used with standard deep-learning machinery to solve various downstream tasks.}
    \label{fig:framework}
\end{figure*}

\section{Introduction}
\label{sec:intro}

Computer vision has always been concerned with understanding the 3D world around us.
One of the main challenges when dealing with 3D data is the representation strategy, which was addressed over the years by introducing various discrete representations, including voxel grids, point clouds, and meshes. 
Each representation has its advantages and disadvantages, especially when it comes to processing it through deep learning, leading to the development of a plethora of ad-hoc algorithms \cite{pointnet++,dgcnn,subdivnet} for each coexisting representation. Hence, no standard way to store and process 3D data has yet emerged.

Recently, a new representation has been proposed, called Neural Fields \cite{neuralfields} (\nf{}s). They are continuous functions defined at all spatial coordinates, parameterized by a neural network such as a Multi-Layer Perceptron (MLP).
In the context of 3D world representation, various types of \nf{}s have been explored. 
Some of the most common \nf{}s utilize the Signed/Unsigned Distance Field (\sdf{}/\udf{}) \cite{deepsdf,ndf,implicitgeoreg,neurallod} and the Occupancy Field (\of{}) \cite{occupancynetworks,convoccnet} to represent the 3D surfaces or volumes of the objects in the scene.
Alternatively, strategies seeking to capture both geometries and appearances often leverage the Radiance Field (\nerf{}), as shown in the pioneering approach NeRF \cite{nerf}.

Representing a 3D scene by encoding it with a continuous function parameterized as an MLP separates the memory cost of the representation from the spatial resolution. In other words, starting from the same fixed number of parameters, it is possible to reconstruct a surface with arbitrarily fine resolution or to render an image with arbitrarily high quality.
Furthermore, the identical neural network architecture can be applied to learn various field functions, offering the possibility of a unified framework for representing 3D objects.

Owing to their efficacy and potential benefits, 3D \nf{}s are garnering growing interest from the scientific community, as evidenced by the frequent publication of novel and impressive results \cite{instant,acorn,neurallod,lip-mlp}. This leads us to speculate that, in the near future, \nf{}s could establish themselves as a standard way to store and communicate 3D data. It is conceivable that repositories hosting digital twins of 3D objects, exclusively realized as MLPs, might become widely accessible.

The above scenario prompts an intriguing research question: can 3D \nf{}s be directly processed using deep learning pipelines for solving downstream tasks, as it is commonly done with discrete representations such as point clouds or images? For instance, is it feasible to classify an object by directly processing the corresponding NeRF without rendering any image from it?

Since \nf{}s are neural networks, there is no straightforward way to process them. 
A recent work in the field, Functa \cite{functa}, fits the whole dataset with a shared network conditioned on a different embedding for each data.
In this formulation, a solution could be to use such embeddings as the input for downstream tasks.
Nevertheless, representing an entire dataset through a shared network poses a formidable learning challenge, as the network encounters difficulties in accurately fitting all the samples (see \cref{sec:comparison_with_recent}). 

On the contrary, recent studies, including SIREN \cite{siren} and others \cite{metasdf, dupont2021coin, strumpler2021implicit, zhang2021implicit, tancik2020fourier}, have demonstrated that it is possible to achieve high-quality reconstructions by tailoring an individual network to each input sample. This holds true even when dealing with complex 3D shapes or images. Furthermore, constructing an individual \nf{} for each object is more adaptable to real-world deployment, as it does not require the availability of the entire dataset to fit each individual data. The increasing popularity of such methodologies suggests that adopting the practice of fitting an individual network is likely to become commonplace in learning \nf{}s.

Therefore, in the former version of this paper \cite{deluigi2023inr2vec}, we explored conducting downstream tasks using deep learning pipelines on 3D data represented as \textbf{individual} \nf{}s. Recently, several methods addressing this topic have been published, such as NFN \cite{zhou2023neural}, NFT \cite{zhou2023neural}, and DWSNet \cite{navon2023equivariant}, and all of them process individual \nf{}s, supporting this paradigm.

Using \nf{}s as input or output data is intrinsically non-trivial, as the MLP of a single \nf{} can encompass hundreds of thousands of parameters.
However, deep models inherently present a significantly redundant parameterization of the underlying function, as shown in \cite{lottery, compression_survey}.
As a result, we explore whether and how an answer to the research question mentioned earlier might be identified within a representation learning framework. We present an approach that encodes individual \nf{}s into compact and meaningful embeddings, making them suitable for diverse downstream tasks. 
We name this framework \algoname{}, shown in \cref{fig:framework}.


Our framework has at its core an encoder designed to produce a task-agnostic embedding representing the input \nf{} by processing only the \nf{} weights.
These embeddings can seamlessly be used in downstream deep learning pipelines as we validate for various tasks, like classification, retrieval, part segmentation, unconditioned generation, completion, and surface reconstruction. 
Remarkably, the last two tasks become achievable by learning a straightforward mapping between the embeddings generated using our framework, as embeddings derived from \nf{}s exist in low-dimensional vector spaces, regardless of the underlying implicit function. For instance, we can learn the mapping between \nf{}s of incomplete objects into \nf{}s of normal ones. Then, we can complete shapes by exploiting this mapping, \eg{} we can map the \nf{} of an airplane with a missing wing into the \nf{} of a complete airplane.
Furthermore, we show that \algoname{} learns a smooth latent space, which enables the interpolation of \nf{}s representing previously unseen 3D objects.

This paper builds on our previous work \cite{deluigi2023inr2vec}, with revisions to the overall framework and thorough experiments on novel scenarios. Specifically, the key differences with \cite{deluigi2023inr2vec} are:

\begin{itemize} 
    \item 
    In \cite{deluigi2023inr2vec}, we focused solely on neural fields representing the surfaces of 3D objects. In this extended version, we also tackle the processing of neural fields capturing objects' geometry and appearance. Specifically, we extend our framework to perform deep learning tasks on NeRFs by directly processing their MLPs weights.
    

    \item The processing of MLPs parametrizing \nf{}s has been investigated in works published contemporaneously or subsequently to \cite{deluigi2023inr2vec}.
    We extend our literature review by including these recent papers, and we evaluate them to foster progress in this emerging topic and facilitate future comparisons.
\end{itemize} 

Overall, the summary of our work contributions is:
\begin{itemize}
    \item We propose and investigate the novel research problem of applying deep learning directly on individual \nf{}s representing 3D objects.
    \item We introduce \algoname{}, a framework designed to derive a meaningful and compact representation of an input \nf{} solely by processing its weights, without the need to sample the underlying function.
    \item We demonstrate that a range of tasks, typically tackled with intricate frameworks tailored to specific representations, can be effectively executed using simple deep learning tools on \nf{}s embedded by \algoname{}, regardless of the signal underlying the \nf{}s.
    \item We demonstrate the versatility of \algoname{} by successfully applying it to neural fields that capture either the geometry alone or the combined information of both geometry and appearance of 3D objects.
    \item We analyze recent methods for processing \nf{}s in terms of classification accuracy and representation quality. We build the first evaluation benchmark for \nf{} classification.
\end{itemize}


Additional details, code, and datasets are available at \url{https://cvlab-unibo.github.io/nf2vec}.

\section{Related Work}
\label{sec:related}

\textbf{Neural fields.}
Recent approaches have shown the ability of MLPs to parameterize fields representing any physical quantity of interest \cite{neuralfields}. 
The works focusing on representing 3D shapes with MLPs rely on fitting functions such as the unsigned distance \cite{ndf}, the signed distance \cite{deepsdf,implicitgeoreg,scenereprnet,jiang2020local,convoccnet, atzmon2020sal}, or the occupancy \cite{occupancynetworks,chen2019learning}.
Among these approaches, sinusoidal representation networks (SIRENs) \cite{siren} use periodical activation functions to capture the high-frequency details of the input data.
In addition to representing shapes, some of these models have been extended to encode object appearance \cite{saito2019pifu,nerf,scenereprnet,oechsle2019texture,niemeyer2020differentiable}, or to include temporal information \cite{niemeyer2019occupancy}.
Among these recent approaches, modeling the radiance field of a scene \cite{nerf} has proven to be the critical factor in obtaining excellent scene representations.
In our work, we employ  \nf{}s encoding \sdf{}, \udf{}, \of{}, and \nerf{} as input data for deep learning pipelines.

\textbf{Deep learning on neural networks.}
Several works have explored using neural networks to process other neural networks. \cite{Unterthiner2020PredictingNN} utilizes a network's weights as input and forecasts its classification accuracy. Another approach \cite{urholt2021selfsupervised} involves learning a network representation through a self-supervised learning strategy applied to the $N$-dimensional weight array. These representations are then employed to predict various characteristics of the input classifier. In contrast, \cite{knyazev2021parameter,jaeckle2021generating,Lu2020Neural} depict neural networks as computational graphs, subsequently processed by a Graph Neural Network (GNN). This GNN is tasked with predicting optimal parameters, adversarial examples, or branching strategies for verifying neural networks.

These works view neural networks as algorithms, primarily focusing on forecasting properties like accuracy. In contrast, some recent studies handle networks that implicitly represent 3D data, thus tackling various tasks directly from their weights, essentially treating neural networks as input/output data.
Functa \cite{functa} tackles this scenario by acquiring priors across the entire dataset using a shared network and subsequently encoding each sample into a concise embedding employed for downstream discriminative and generative tasks. We note that in this formulation, each neural field is parametrized by both the shared network and the embedding.
It is worth pointing out that, though not originally proposed as a framework to process neural fields, DeepSDF \cite{deepsdf} learns dataset priors by optimizing a reconstruction objective through a shared auto-decoder network conditioned on a shape-specific embedding. Thus, the embeddings learned by DeepSDF may be used for neural processing tasks, as done in Functa.

\begin{figure*}[t]
    \centering
    \includegraphics[width=0.9\textwidth]{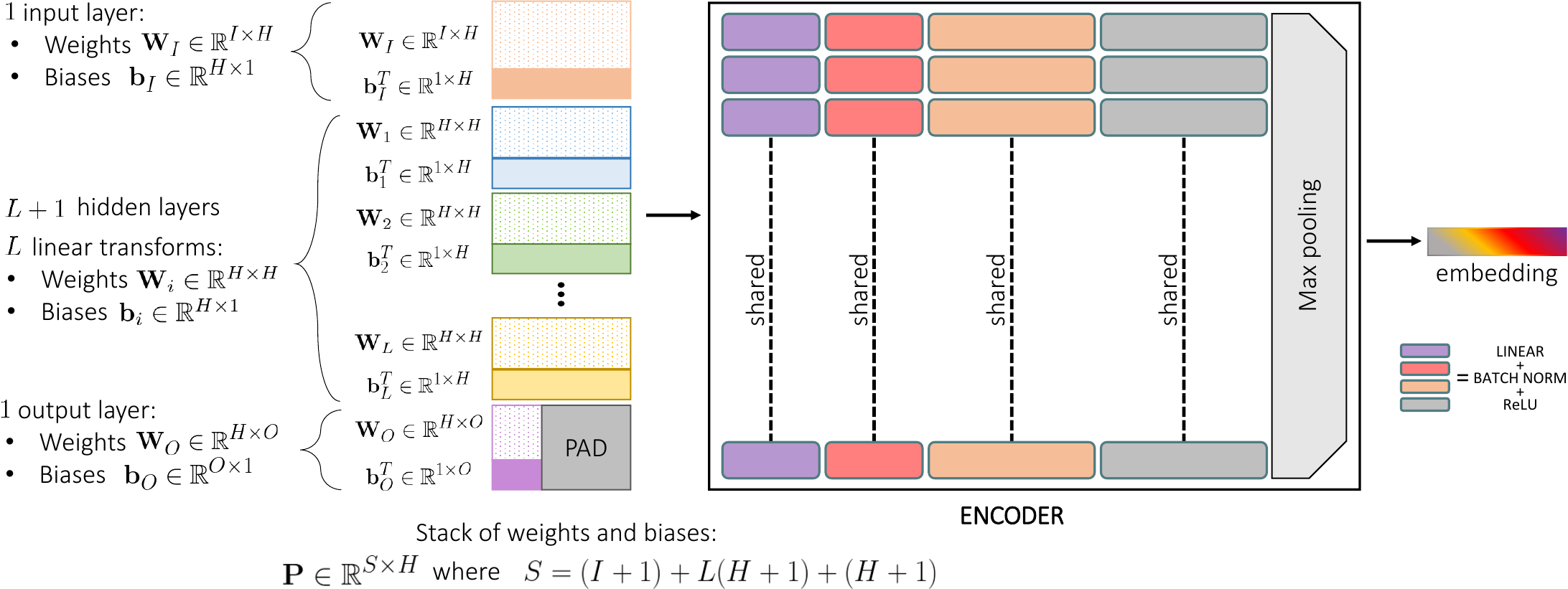}
    \caption{\textbf{Encoder architecture.}
    \textbf{Left}: Given a \nf{}, we stack its weights and biases to form a matrix $\mathbf{P}$. \textbf{Right}: The \algoname{} encoder is a series of linear layers with batch-norms and ReLU activation functions. It processes each row of $\mathbf{P}$ independently and then aggregates all rows of one \nf{} with a max-pooling to produce a compact embedding employed for downstream tasks.}
    \label{fig:encoder}
\end{figure*}

\begin{table}[t]
    \centering
    \begin{tabular}{ccccc}
        \toprule
        \multicolumn{3}{c}{\nf} & Our encoder & MLP encoder \\
        \cmidrule(lr){1-3}\cmidrule(lr){4-4}\cmidrule(lr){5-5}
        Hidden dim. & \#Layers & \#Params & \#Params & \#Params \\
        \midrule
        512 & 4 & $\sim$800K & $\sim$3M & $\sim$800M \\
        512 & 8 & $\sim$2M & $\sim$3M & $\sim$2B \\
        512 & 16 & $\sim$4M & $\sim$3M & $\sim$4B \\
        1024 & 4 & $\sim$3M & $\sim$3.5M & $\sim$3B \\
        1024 & 8 & $\sim$7M & $\sim$3.5M & $\sim$7.5B \\
        1024 & 16 & $\sim$15M & $\sim$3.5M & $\sim$16B \\
        \bottomrule
    \end{tabular}
    \caption{\textbf{Num. of parameters of our vs an MLP encoder.}}
    \label{tab:encoder_num_params}
\end{table}

However, shared network frameworks face several challenges, as they struggle to reconstruct the underlying signal with high fidelity and require an entire dataset to learn the neural field of an object.
In response, recent approaches have shifted their focus on processing \nf{}s learned on individual data, \eg{} a specific object or scene.
The first framework adopting this view was proposed in our previous paper version \cite{deluigi2023inr2vec}. This approach leverages representation learning to condense individual \nf{}s of 3D shapes into embeddings, serving as input for subsequent tasks.
\cite{ballerini2024clip2nerf} has recently built upon this idea to learn a bidirectional mapping between image/text and NeRF latent spaces.
Recognizing that MLPs exhibit weight space symmetries \cite{hecht1990algebraic}, where hidden neurons can be permuted across layers without altering the network's function, recent approaches such as DWSNet \cite{navon2023equivariant}, NFN \cite{zhou2023permutation}, and NFT \cite{zhou2023neural} leverage these symmetries as an inductive bias to create innovative architectures tailored for MLPs. DWSNet and NFN design neural layers equivariant to the permutations inherent in MLPs. In contrast, NFT achieves permutation equivariance by removing the positional encoding from a Transformer architecture. A recent work by \cite{cardace2024neural} overcomes the
need to deal with MLP symmetries by proposing a Transformer-based architecture that processes \nf{}s with tri-planar grid features by focusing on those discrete features only.

Recently, HyperDiffusion \cite{hyperdiffusion} has proposed a generative diffusion approach to synthesize \nf{} parameters. Like us, it employs MLPs optimized to represent individual data.

\section{Learning to Represent \nf{}s}
\label{sec:learning_to_repr}

This paper explores the possibility and the methodology of directly utilizing \nf{}s for downstream tasks. Specifically, can we classify an object implicitly encoded in a \nf{}, and if so, how?
As outlined in \cref{sec:intro}, we condense the redundant information encoded in the weights of \nf{}s into latent codes by a representation learning framework. These codes can then be efficiently processed using standard deep-learning pipelines.
Our framework, dubbed \algoname{}, comprises an encoder and a decoder. In the following sections, we first provide some basic knowledge about what a 3D neural field is, then we deepen the reasons behind the architectural choices for both components and describe the representation learning protocol.

\textbf{3D Neural Fields.} A field is a physical quantity defined for all domain coordinates. We focus on fields describing the 3D world, thus operating on $\mathbb{R}^3$ coordinates $\mathbf{p} = (x, y, z)$. We consider the 3D fields commonly used in computer vision and graphics, such as the \sdf{} \cite{deepsdf} and \udf{} \cite{ndf}, which map coordinates to the signed and unsigned distance from the closest surface, the \of{} \cite{occupancynetworks}, which computes the occupancy probability of each position, and the \nerf{} \cite{nerf}, that outputs $(r, g, b)$ colors and density $\sigma$ for each 3D point.
A field can be modeled by a function, $f$, parameterized by $\theta$. Thus, for any point $\mathbf{p}$, the field is given by $f(\mathbf{p})$.
If parameters $\theta$ are the weights of a neural network, $f$ is said to be a Neural Field  (\nf{}) \cite{neuralfields}.



\begin{figure*}[t]
    \centering
    \includegraphics[width=\textwidth]{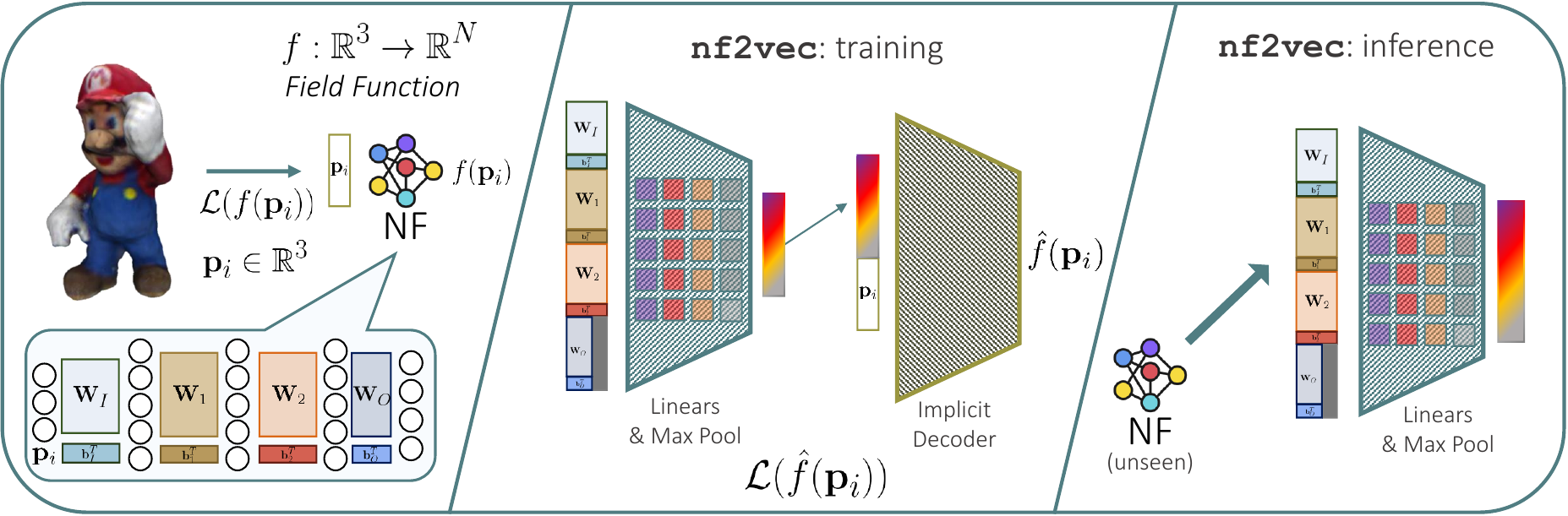}
    \caption{\textbf{Training and inference of \algoname{}.} 
    \textbf{Left:} A Neural Field (\nf{}) represents a 3D object. The \nf{} is composed of an MLP that parametrizes a field function $f$.
    \textbf{Center:} \algoname{} encoder is trained together with an implicit decoder. The implicit decoder processes the embedding produced by the encoder to estimate field values $\hat{f}$. We train the framework similarly to the input \nf{}.
    \textbf{Right:} At inference time, the learned encoder can be used to obtain a compact embedding from unseen \nf{}s.}
    \label{fig:inr2vec}
\end{figure*}

\textbf{Encoder.}
The encoder takes as input the weights of a \nf{} and produces a compact embedding that encodes all the relevant information of the input \nf{}.
Designing an encoder for \nf{}s poses a challenge in handling weights efficiently to avoid excessive memory usage. While a straightforward solution might involve using an MLP encoder to map flattened weight vectors to desired dimensions, this approach becomes impractical for larger \nf{}s. For instance, given a 4-layer 512-neurons \nf{}, mapping its 800K parameters to a 1024-sized embedding space would require an encoder with roughly 800M parameters, making this approach prohibitive.
Thus, we focus on developing an encoder architecture that scales gracefully with the size of the input \nf{}.

Following conventional practice \cite{siren,metasdf,dupont2021coin,strumpler2021implicit,zhang2021implicit}, we consider \nf{}s composed of an MLP with several hidden layers, each with $H$ nodes.
The linear transformation between two consecutive hidden layers is parameterized by a matrix of weights $\mathbf{W}_{i} \in \mathbb{R}^{H \times H}$ and a vector of biases $\mathbf{b}_{i} \in \mathbb{R}^{H \times 1}$. Thus, stacking $\mathbf{W}_{i}$ and ${\mathbf{b}_{i}}^T$, the mapping between two consecutive layers can be represented by a single matrix $\mathbf{P}_i \in \mathbb{R}^{(H+1) \times H}$.
Additionally, an MLP features also an input layer, parametrized by $\mathbf{W}_{I} \in \mathbb{R}^{I \times H}$ and a vector of biases $\mathbf{b}_{I} \in \mathbb{R}^{H \times 1}$, and an output layer, parametrized by $\mathbf{W}_{O} \in \mathbb{R}^{H \times O}$ and a vector of biases $\mathbf{b}_{O} \in \mathbb{R}^{O \times 1}$. The input and output layers can be represented by two matrices, $\mathbf{P}_I \in \mathbb{R}^{(I+1) \times H}$ and $\mathbf{P}_O \in \mathbb{R}^{(H+1) \times O}$.
Considering that for a \nf{} with $L + 1$ hidden layers there are $L$ linear transformations between them,
we can store all the weights of the \nf{} by stacking the input matrix $\mathbf{P}_I$ with all the $L$ matrices $\mathbf{P}_i$ and with the output matrix $\mathbf{P}_O$.
As $\mathbf{P}_O$ has a different number of columns ($O$), we pad it with zeros before stacking.
The final stacked matrix $\mathbf{P}$ has dimension $S \times H$ -- where $S = (I + 1) + L(H+1) + (H + 1)$ -- and represents the input for our encoder, as shown in the left part of \cref{fig:encoder}.

\algoname{} encoder consists of a series of linear layers with batch normalization and ReLU non-linearity followed by final max pooling. At each stage, the input matrix $\mathbf{P}$ is transformed by one linear layer, processing each row independently.
The final max pooling compresses all the rows into a single one, obtaining the desired embedding. An architecture overview is depicted in \cref{fig:encoder}. 

Our proposed architecture scales gracefully to bigger input \nf{}s as supported by the analysis in \cref{tab:encoder_num_params}, that reports the parameters of our encoder \wrt{} those of a generic MLP encoder while varying the input \nf{} dimension.


It is worth observing that the randomness involved in fitting an individual \nf{} (weights initialization, data shuffling, etc.) causes the weights in the same position in the \nf{} architecture not to share the same role across \nf{}s. Thus, \algoname{} encoder would have to deal with input vectors whose elements capture different information across the different data samples, making it impossible to train the framework.
However, the use of a shared, pre-computed initialization has been advocated as a good practice when fitting \nf{}s, \eg{} to reduce training time by means of meta-learned initialization vectors, as done in MetaSDF \cite{metasdf} and in Functa \cite{functa}, or to obtain desirable geometric properties \cite{implicitgeoreg}.
We empirically found that following such a practice, \ie{} initializing all \nf{}s with the same random vector, favors the alignment of weights across \nf{}s and enables the convergence of our framework. More analysis can be found in \cref{sec:same_init}.

\textbf{Decoder.}
When learning to encode \nf{}s, we are interested in storing the information about the represented object rather than the values of the input weights.
Therefore, the adopted decoder predicts the original field values rather than reconstructing the input weights in an auto-encoder fashion.
In particular, during training, we adopt an implicit decoder inspired by \cite{deepsdf}, which takes in input the embeddings produced by the encoder and a spatial coordinate $\mathbf{p}_i$ and decodes the original field values (see \cref{fig:inr2vec} center). We denote as $\hat{f}(\mathbf{p}_i)$ the predicted field value.

\begin{figure}
    \centering
    \includegraphics[width=\linewidth]{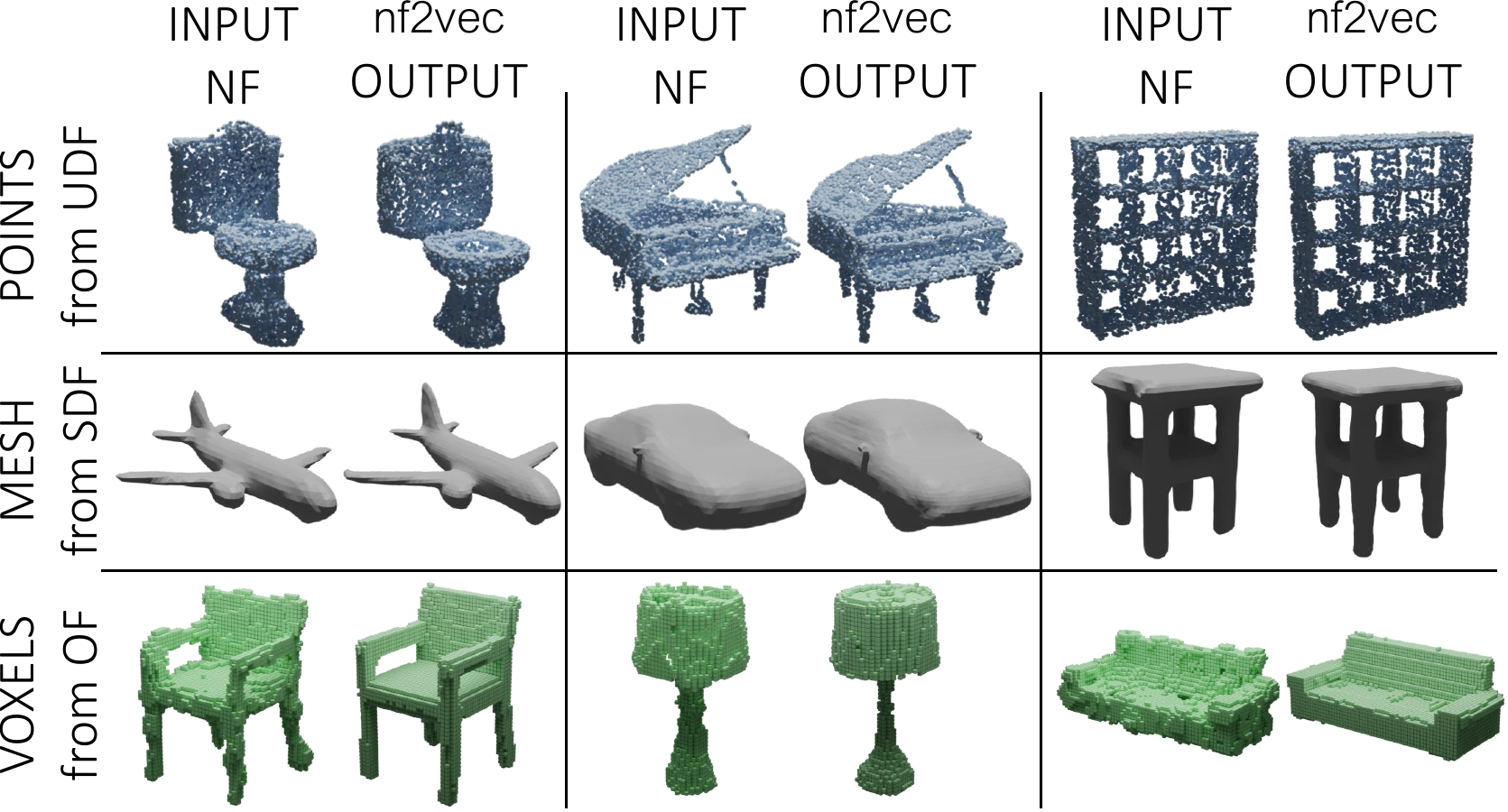}
    \caption{\textbf{\inrtovec{} reconstructions.}}
    \label{fig:qualitatives}
\end{figure}
\begin{figure}
    \centering
    \includegraphics[width=1.0\linewidth]{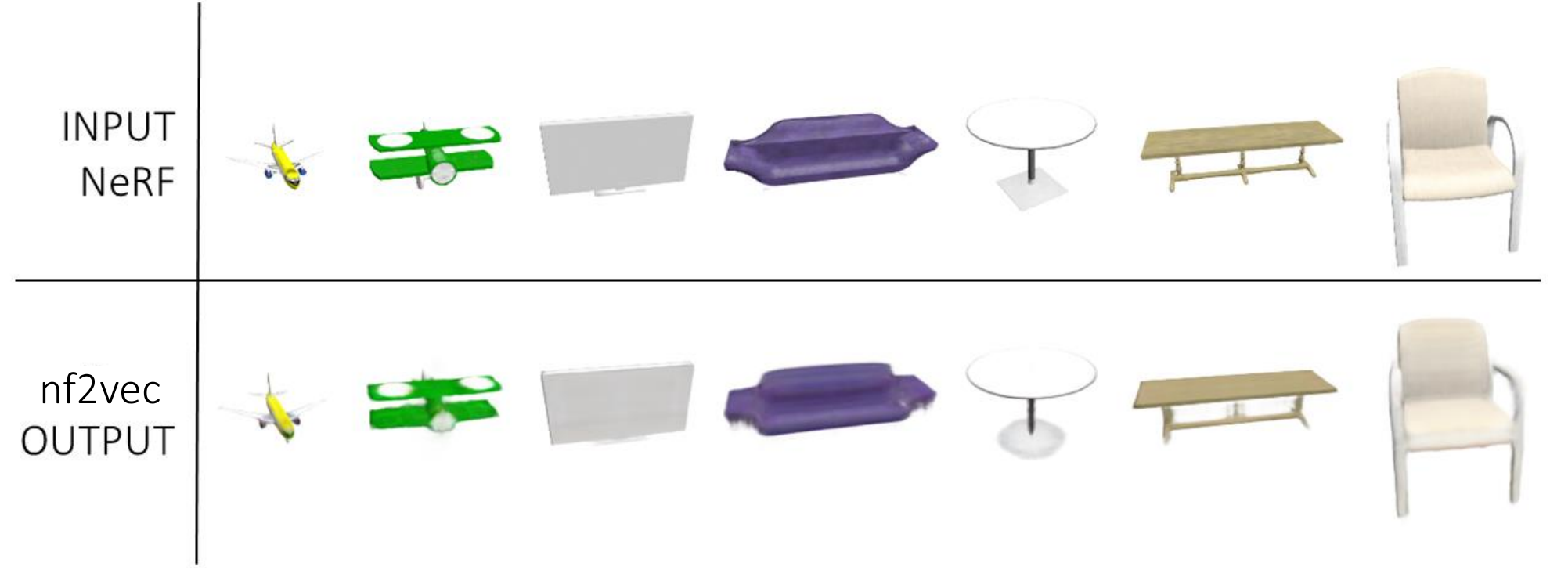}
    \caption{\textbf{\nerfalgoname{} reconstructions.}}
    \label{fig:nerf_reconstructions}
\end{figure}

\textbf{Training.}
We train our encoder and decoder following the input \nf{}s training strategy.
For instance, when dealing with \udf{}, \sdf{}, and \of{} representing 3D surfaces, we supervise the framework directly using the ground truth field values computed from point clouds, voxel grids, or triangle meshes representing those surfaces.
Differently, when processing NeRFs
we employ volumetric rendering \cite{nerf} on the radiance field values predicted by the decoder to obtain the RGB intensities of image pixels, and we supervise the framework directly with a regression loss between predicted and true RGB values.

To better understand the procedure, let us take the example where we aim to learn to represent \udf{}s.
We create a set of 3D queries paired with the values of the \udf{} at those locations. The decoder takes in input the embedding produced by the encoder concatenated with the 3D coordinates of a query point and estimates the \udf{} for this location. The whole encoder-decoder is supervised to minimize the discrepancy between the estimated and correct \udf{} values.


\textbf{Inference.}
After the overall framework has been trained end to end, the frozen encoder can be used to compute embeddings of unseen \nf{}s with a single forward pass (see \cref{fig:inr2vec} right) while the implicit decoder can be used, if needed, to reconstruct the discrete representation given an embedding.
\textit{We highlight that no discrete representations are required at inference time.}\\

The presented \algoname{} framework shares the same high-level structure of the originally proposed approach {\tt inr2vec} \cite{deluigi2023inr2vec}. However, as shown in \cref{sec:experiments_nerf}, in this extended work, we show that our framework can be applied to more complex neural fields such as NeRFs. Thus, we dub it \algoname{} to emphasize its generality.


\begin{figure}[t]
    \centering
    \includegraphics[width=\linewidth]{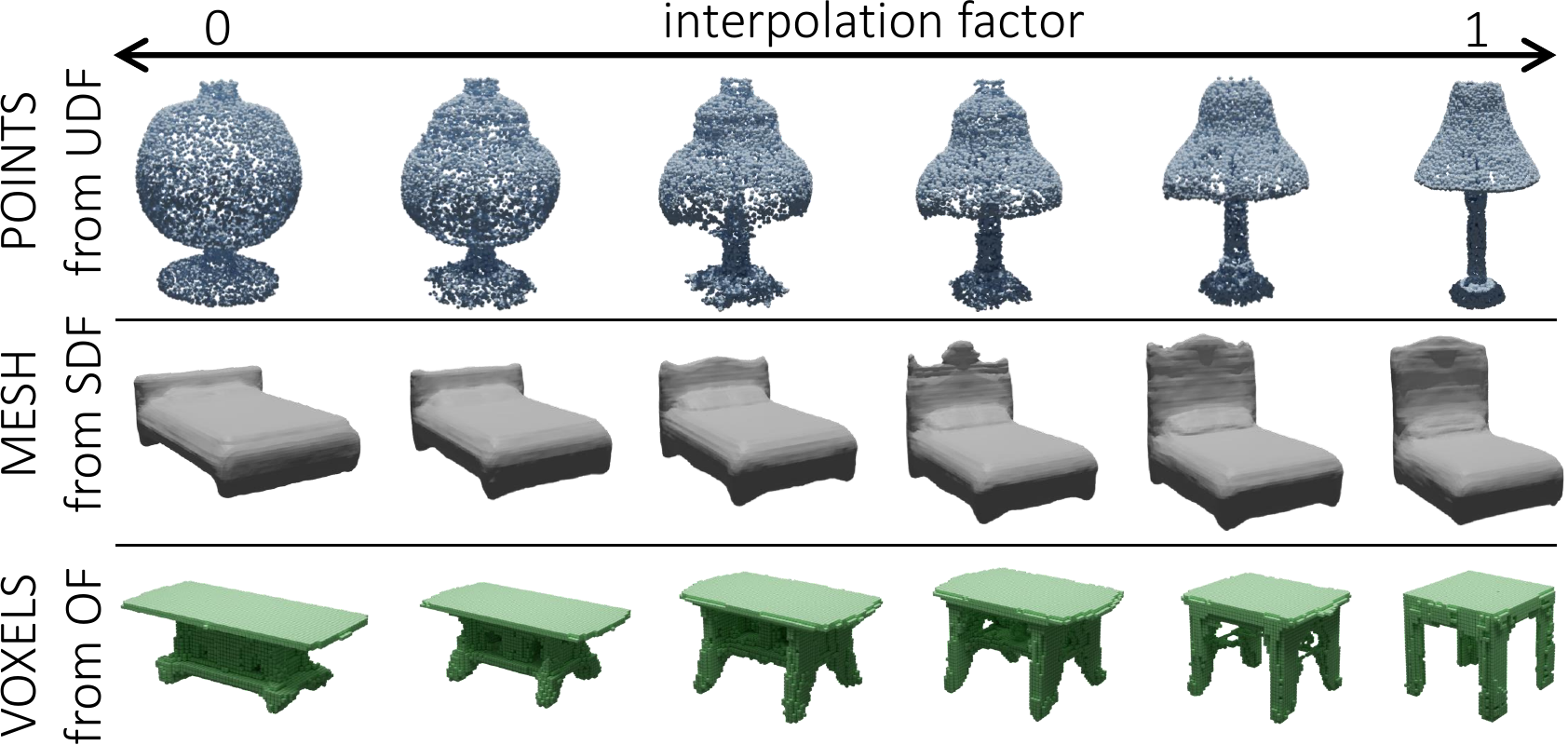}
    \caption{\textbf{\inrtovec{} latent space interpolation.}}
    \label{fig:interpolations}
\end{figure}
\begin{figure}[t]
    \centering
    \includegraphics[width=1.0\linewidth]{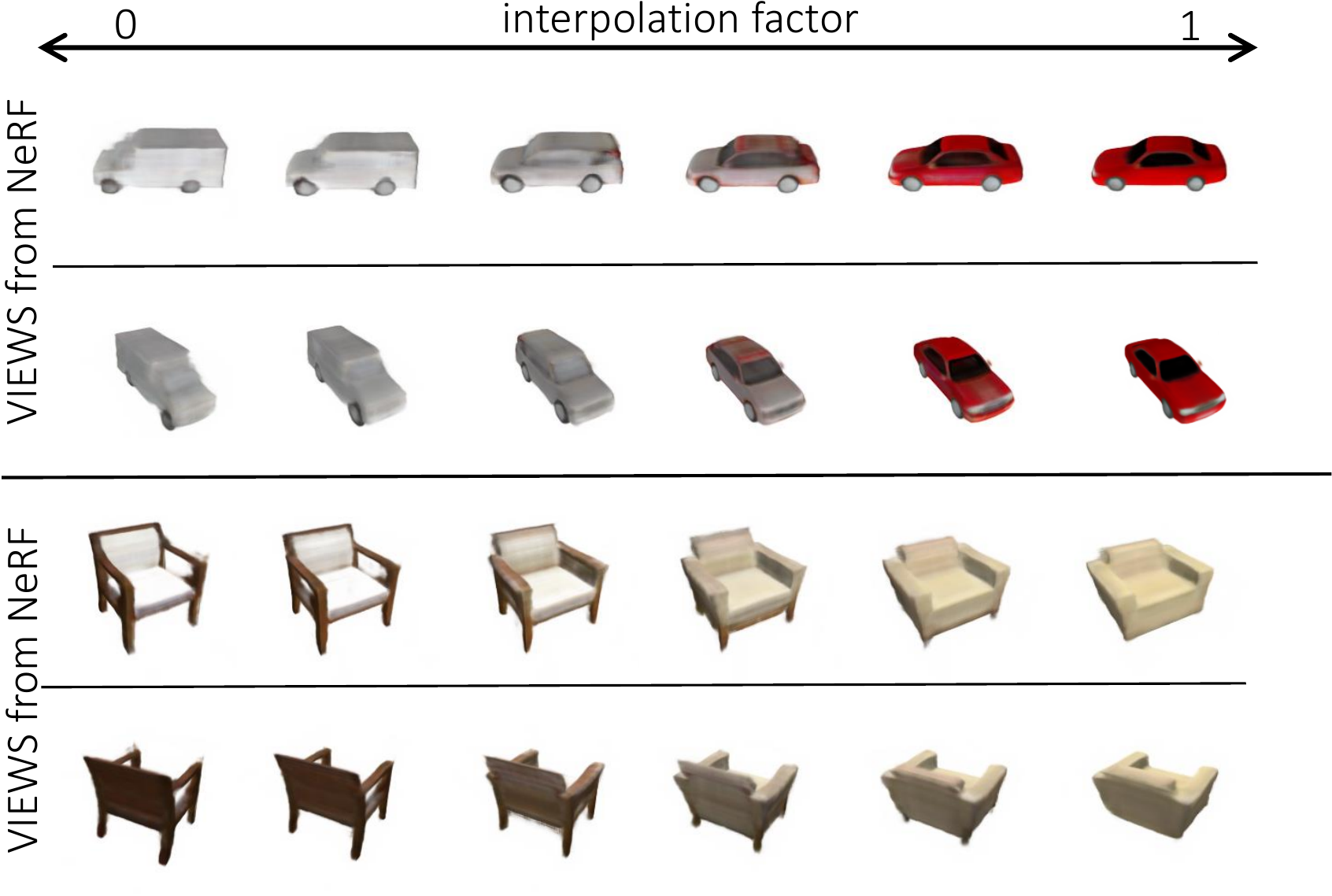}
    \caption{\textbf{\nerfalgoname{} latent space interpolation.}}
    \label{fig:nerf_interpolations}
\end{figure}
\begin{figure}[t]
    \centering
    \includegraphics[width=1.0\linewidth]{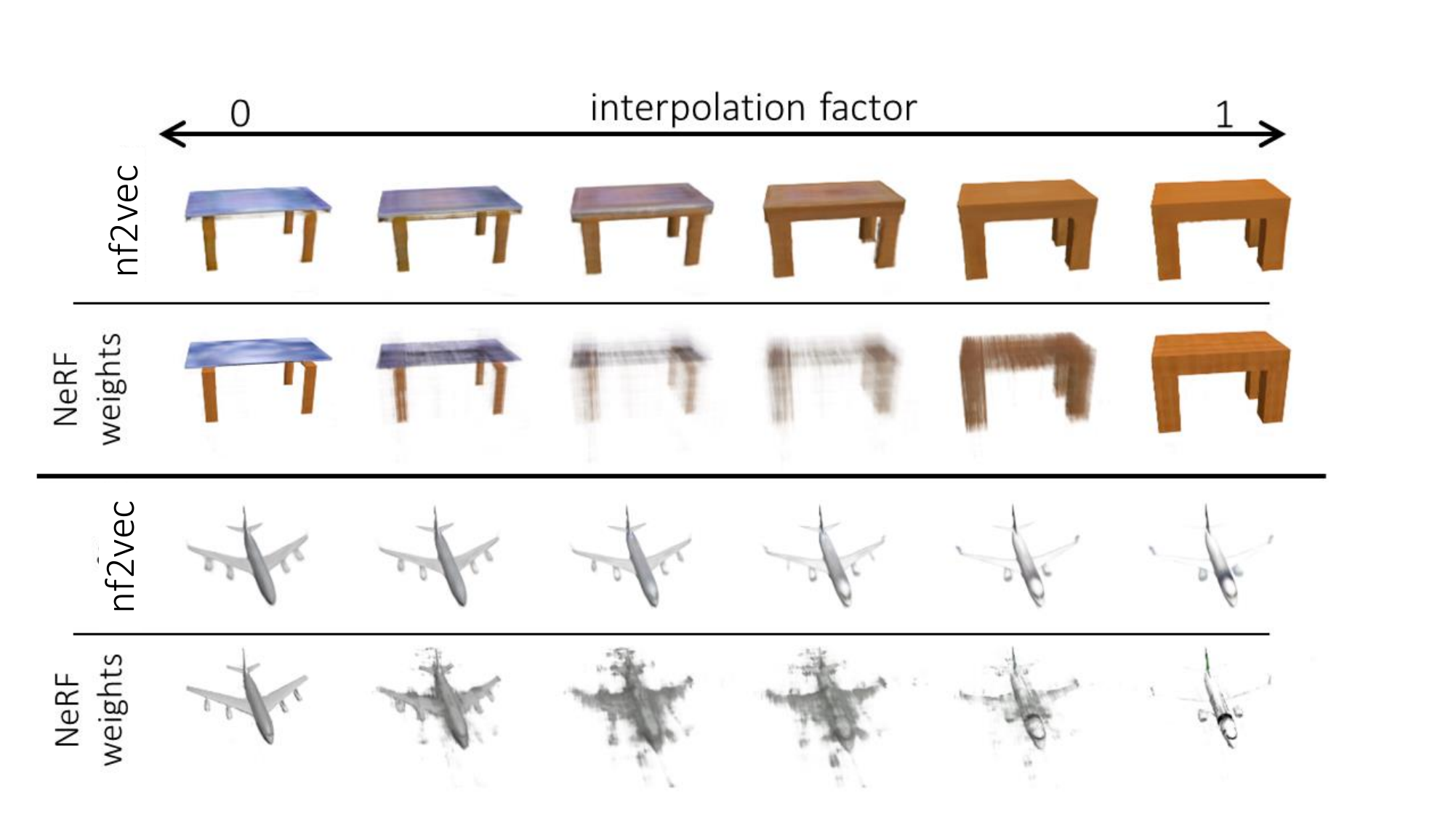}
    \caption{\textbf{\nerfalgoname{} latent space interpolation vs NeRF weights interpolation.}}
    \label{fig:nerf_interpolations_baseline}
\end{figure}

\section{Latent Space Properties}

We train \algoname{} on \nf{}s learned from various 3D discrete representations of ShapeNet \cite{chang2015shapenet} surfaces. We use either \udf{} learned from point clouds, \sdf{} learned from meshes, or \of{} learned from voxel grids.
We also train \algoname{} on NeRFs trained on multi-view renderings of ShapeNet \cite{chang2015shapenet} objects.
The following sections explore \algoname{} latent space properties. We investigate the ability to reconstruct the original discrete representation from the compact latent codes extracted with our encoder, the smoothness of the embedding space, and its structure.

\textbf{Reconstruction.}
In \cref{fig:qualitatives}, we compare 3D shapes reconstructed from \nf{}s unseen during training with those reconstructed by the \algoname{} decoder starting from the latent codes yielded by the encoder. We visualize point clouds with 8192 points, meshes reconstructed by marching cubes \cite{marching} from a grid with resolution $128^3$ and voxels with resolution $64^3$. 
Moreover, we conduct the same experiment using as input unseen NeRFs. In \cref{fig:nerf_reconstructions}, we show the images rendered by \algoname{} decoder and those rendered from the input NeRF at $224 \times 224$ resolution.
Though our embedding is dramatically more compact than the original \nf{}, the reconstructed discrete data resembles those of the original input \nf{}.

\textbf{Interpolation.}
In \cref{fig:interpolations} and \cref{fig:nerf_interpolations}, we linearly interpolate between two object embeddings produced by \algoname{}.
Results highlight that the learned latent spaces enable smooth interpolations between shapes represented as \nf{}s.
Notably, in \cref{fig:nerf_interpolations}, color and shapes change smoothly when interpolating two NeRF embeddings. Moreover, the \textit{interpolated} object has an underlying 3D consistency, as visible when rendering it from different camera viewpoints.
Thus, these results demonstrate the capability to generate new plausible shapes and even realistic \nerf{}s by means of simple linear interpolation in \algoname{} latent space.

Additionally, given two input NeRFs, we render images from networks obtained by interpolating their weights. In \cref{fig:nerf_interpolations_baseline}, we compare these results with those obtained from the interpolation of \nerfalgoname{} embeddings. Notably, our interpolations exhibit superior quality. In particular, renderings obtained by averaging the weights of the two NeRFs (interpolation factor 0.5) appear blurred and lack 3D structure. In contrast, renderings produced by \nerfalgoname{} preserve details and maintain 3D consistency.
Our results highlight the efficacy of our representation learning approach in transforming an initially disorganized input weight space into a well-organized latent space.

\begin{figure}
    \centering
    \setlength{\tabcolsep}{1pt}
    \begin{tabular}{ccc}
         \includegraphics[width=0.3\linewidth]{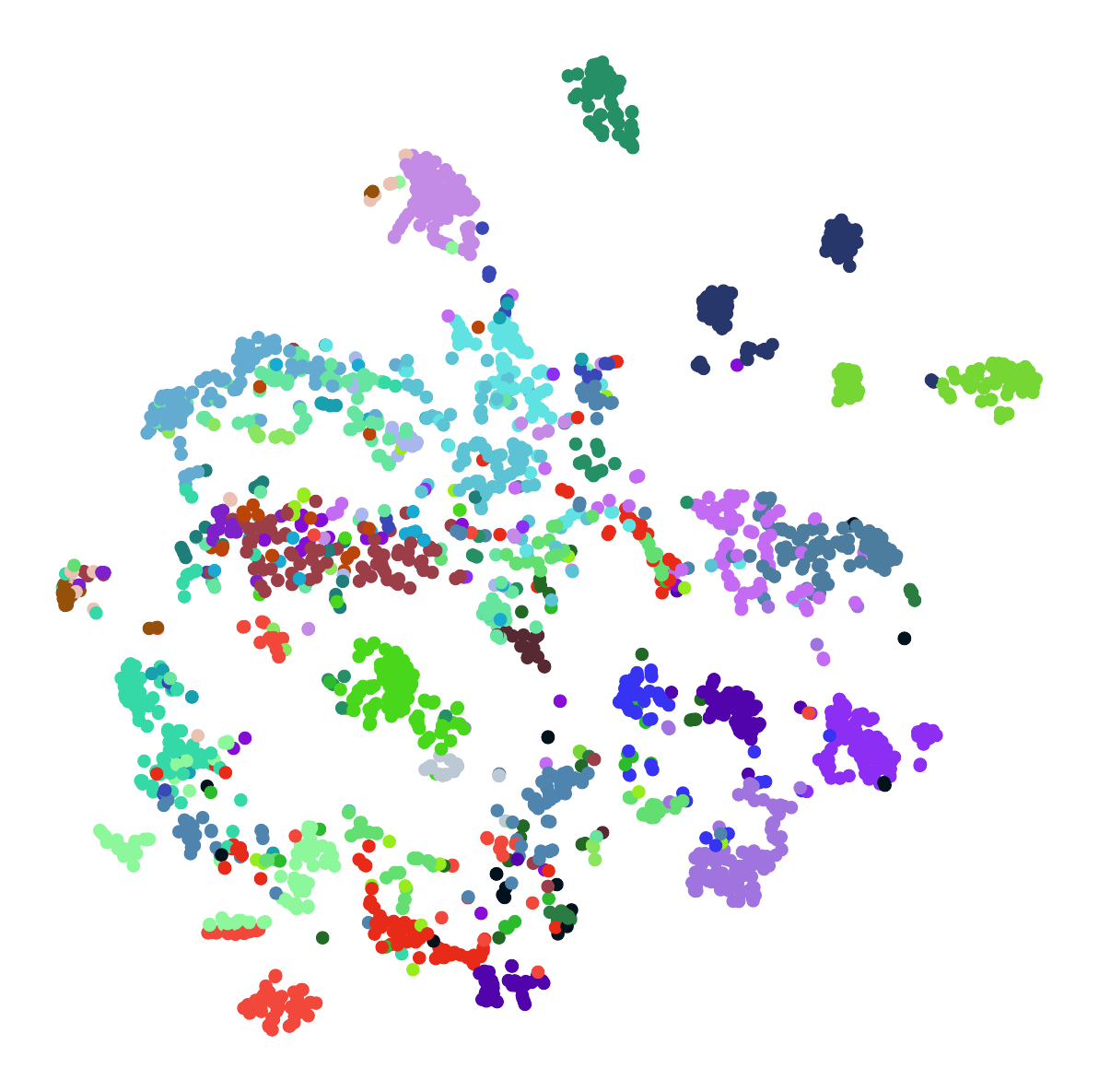} &
         \includegraphics[width=0.3\linewidth]{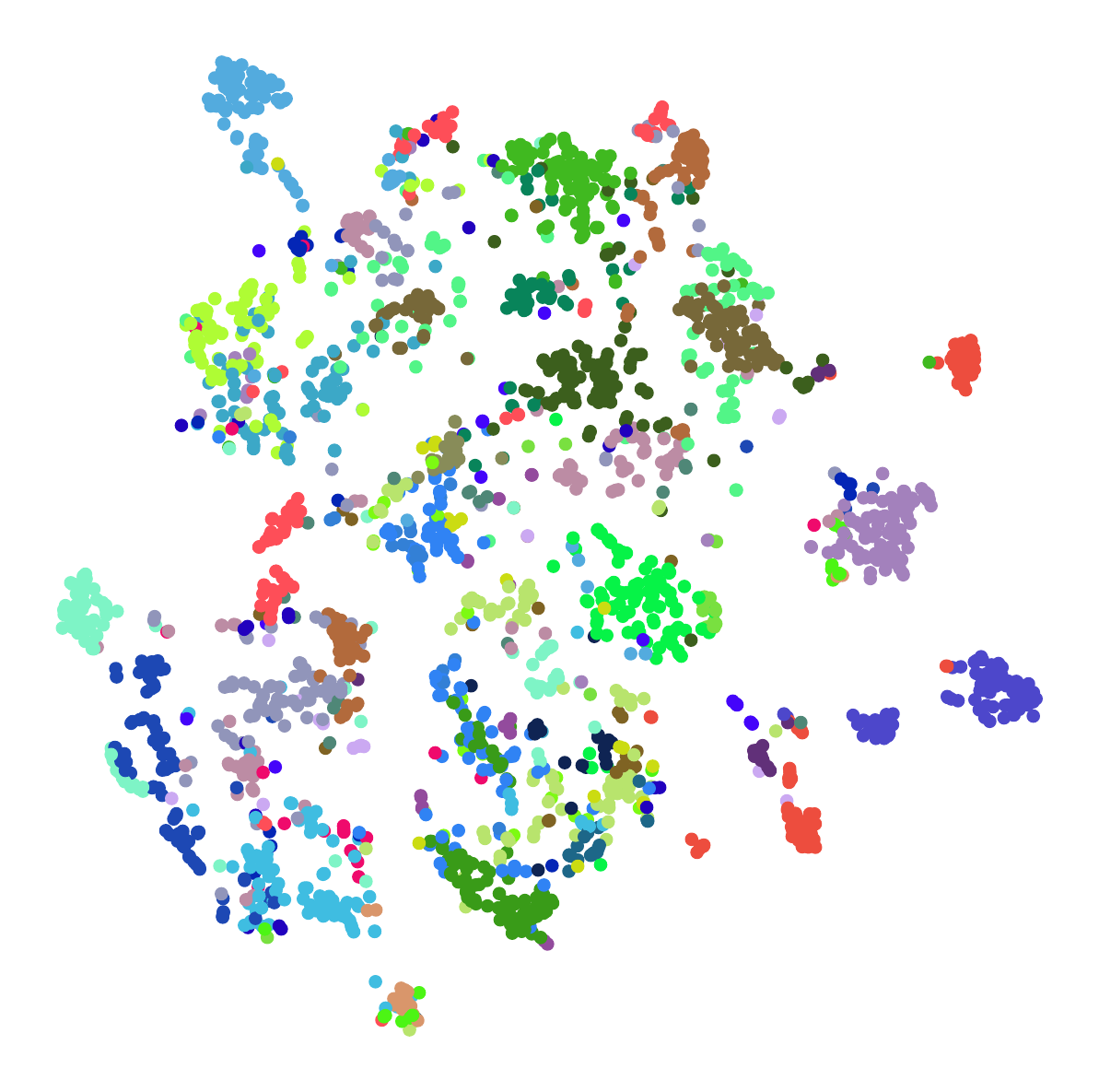} &
         \includegraphics[width=0.3\linewidth]{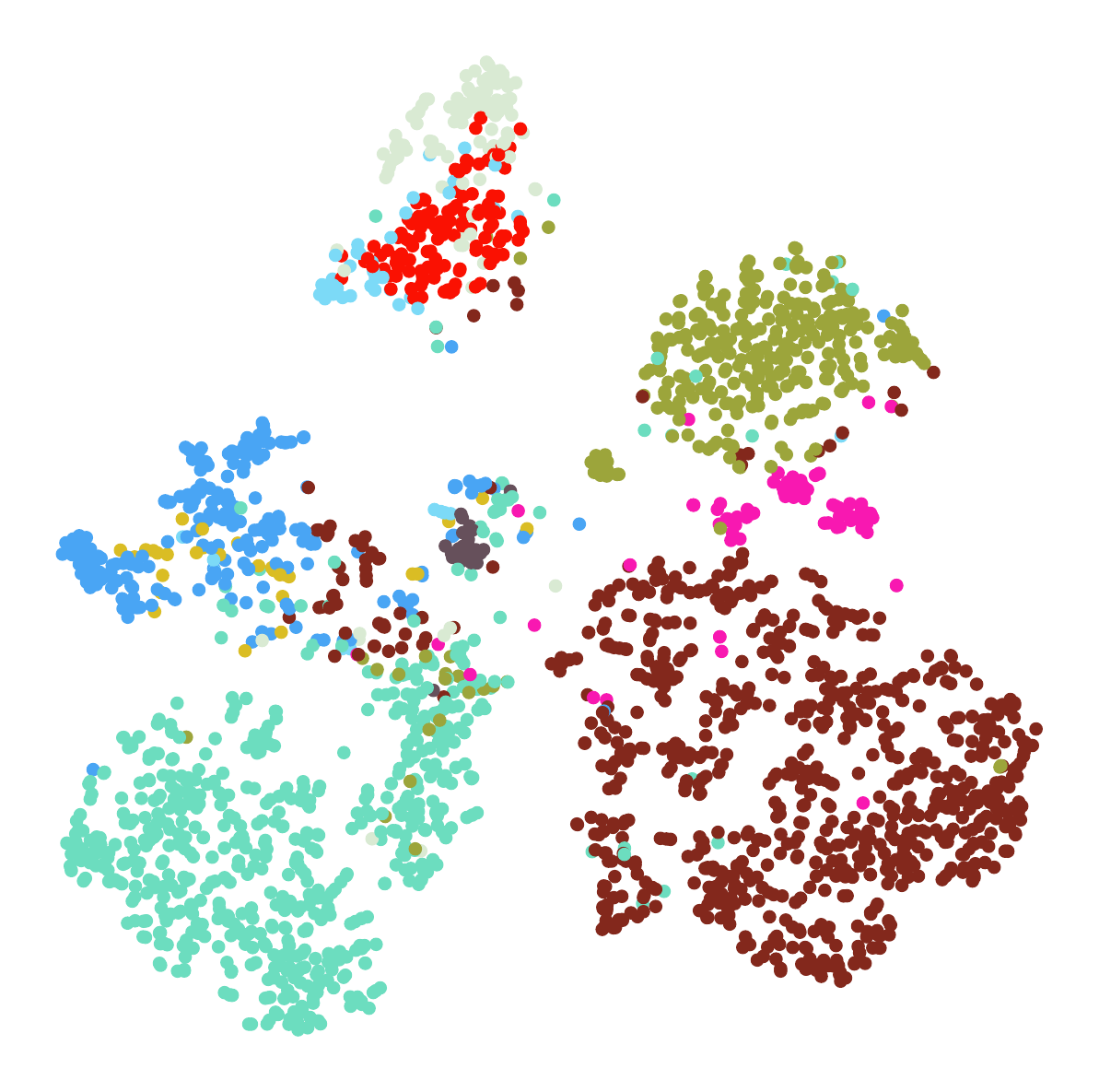} \\
         ModelNet40 &
         Manifold40  & 
         ShapeNet10 \\
         (\udf{}) &
         (\sdf{}) &  
         (\of{})\\
    \end{tabular}
    \caption{\textbf{t-SNE visualizations of \algoname{} latent spaces.} We plot the t-SNE components of the embeddings produced by \algoname{} on the test sets of three datasets, ModelNet40 (left), Manifold40 (center) and Shapenet10 (right). Colors represent the different classes of the datasets.}
\label{fig:tsne}
\end{figure}
\textbf{t-SNE visualization of the latent space.}
In \cref{fig:tsne}, we provide the t-SNE visualization of the embeddings produced by \inrtovec{} when presented with unseen \nf{} of three different datasets: ModelNet40 (\udf{} learned from point clouds), Manifold 40 (\sdf{} learned from meshes), and ShapeNet10 (\of{} learned from voxel grids).
During the training of our framework, the supervisory signal employed does not impose any particular constraints on the organization of the learned latent space. This lack of constraints was deliberate, as it was not necessary for our primary goal -- performing downstream tasks with the generated embeddings. Nevertheless, it is intriguing to note from the t-SNE plots that our algorithm naturally arranges the embeddings in the latent space with a semantic structure, with items of the same category consistently mapped to close positions. This is evident in the colors representing different classes within the datasets under consideration.

\section{Deep Learning on 3D Shapes}
\label{sec:experiments}

This section shows how several tasks dealing with 3D shapes can be tackled by working only with \inrtovec{} embeddings as input and/or output.

\textbf{General settings.}
In all the experiments reported in this section, we convert 3D discrete representations into \nf{}s featuring 4 hidden layers with 512 nodes each, using the SIREN activation function \cite{siren}.
We discard the input and output layers of SIREN MLPs when processing them with \inrtovec{}. This is based on the observation from the earlier version of this paper \cite{deluigi2023inr2vec} that these layers do not provide information beneficial for downstream tasks when using SIREN MLPs as architecture for \nf{}.
We train \inrtovec{} using an encoder composed of four linear layers with respectively 512, 512, 1024, and 1024 features, embeddings with 1024 values, and an implicit decoder with 5 hidden layers with 512 features. In all the experiments, the baselines are trained using standard data augmentation (random scaling and point-wise jittering), while we train both \inrtovec{} and the downstream task-specific networks on datasets augmented offline with the same transformations.

\begin{figure}
    \centering
    \includegraphics[width=0.45\textwidth]{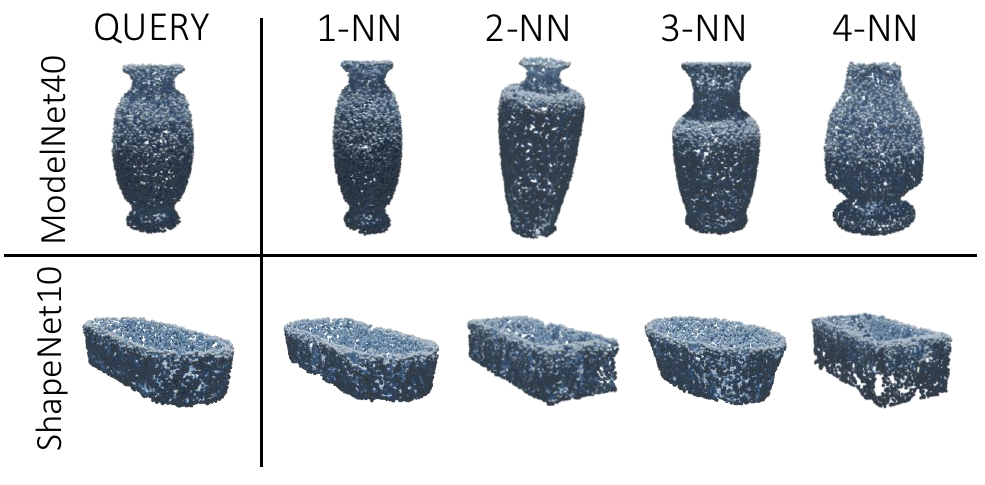}%
    \caption{\textbf{Point cloud retrieval qualitative results.} Given the \inrtovec{} embedding of a query shape, we show the shapes reconstructed from the closest embeddings (L2 distance).
    }
    \label{fig:retrieval}
\end{figure}

\begin{figure}
    \centering
    \includegraphics[width=\linewidth]{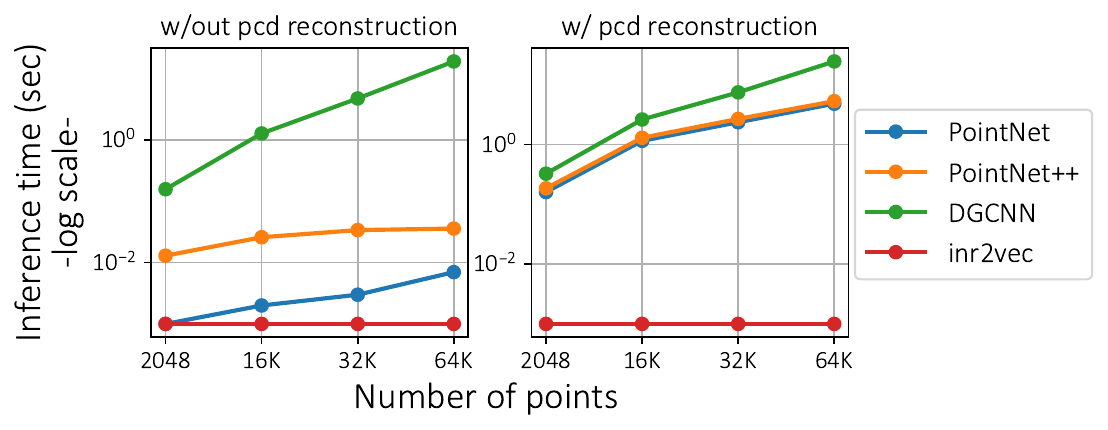}
    \caption{\textbf{Time required to classify \nf{}s encoding $\mathbf{udf}$.} We plot the inference time of standard baselines and of our method, both considering the case in which discrete point clouds are available (left) and the one where point clouds must be reconstructed from the input \nf{}s (right).}
\label{fig:cls_inference_time_details}
\end{figure}

\textbf{Point cloud retrieval.}
We examine the potential of using \inrtovec{} embeddings for representation learning tasks, with 3D retrieval as our benchmark.
We follow the procedure introduced in \cite{chang2015shapenet}, using the Euclidean distance to measure the similarity between embeddings of unseen point clouds from the test sets of ModelNet40 \cite{modelnet} and ShapeNet10 (a subset of 10 classes of the popular ShapeNet dataset \cite{chang2015shapenet}).
For each embedded shape, we select its $k$-nearest-neighbours and compute a Precision Score comparing the classes of the query and the retrieved shapes, reporting the mean Average Precision for different $k$ (mAP@$k$).
Beside \inrtovec{}, we consider three baselines to embed point clouds, which are obtained by training the PointNet \cite{pointnet}, PointNet++ \cite{pointnet++} and DGCNN \cite{dgcnn} encoders in combination with a fully connected decoder similar to that proposed in \cite{fan2017point} to reconstruct the input cloud.
The quantitative findings in \cref{tab:retrieval} reveal that \inrtovec{} not only matches but sometimes exceeds the performance of other baselines, with an average gap of 1.8 mAP compared to PointNet++.
Furthermore, as depicted in \cref{fig:retrieval}, it is evident that the retrieved shapes not only belong to the same class as the query but also exhibit similar coarse structures.
These results highlight that the pretext task used to learn \inrtovec{} embeddings allows encoding relevant shape information.

\begin{table*}
    \centering
    \begin{tabular}{llccccccccc}
        \toprule
        && \multicolumn{3}{c}{ModelNet40}   & \multicolumn{3}{c}{ShapeNet10} & \multicolumn{3}{c}{ScanNet10}  \\
        \cmidrule(lr){3-5}\cmidrule(lr){6-8}\cmidrule(lr){9-11}
        Method & Input & mAP@1 & mAP@5 & mAP@10 & mAP@1 & mAP@5 & mAP@10 & mAP@1 & mAP@5 & mAP@10\\
        \midrule
        PointNet~\cite{pointnet} & Point cloud & 80.1 & 91.7 & 94.4 & 90.6 & 96.6 & 98.1 & 65.7 & 86.2 & 92.6\\ 
        PointNet++~\cite{pointnet++} &  Point cloud & \textbf{85.1} & \textbf{93.9} & \textbf{96.0} & \textbf{92.2} & \textbf{97.5} & \textbf{98.6} & \textbf{71.6} & \textbf{89.3} & 93.7\\
        DGCNN~\cite{dgcnn} &  Point cloud & 83.2 & 92.7 & 95.1 & 91.0 & 96.7 & 98.2 & 66.1 & 88.0 & 93.1\\
        \cmidrule(lr){1-11}
        \inrtovec{} &  \nf{} & 81.7 & 92.6 & 95.1 & 90.6 & 96.7 & 98.1 & 65.2 & 87.5 & \textbf{94.0} \\
        \bottomrule
    \end{tabular}
    \caption{\textbf{Point cloud retrieval quantitative results.}}
    \label{tab:retrieval}
\end{table*}

\begin{table*}
    \centering
    \begin{tabular}{llccccc}
        \toprule
        Method & Input & ModelNet40 & ShapeNet10 & ScanNet10 & Manifold40 & ShapeNet10 \\
        \midrule
        PointNet~\cite{pointnet} &  Point cloud & 88.8 & 94.3 & 72.7 & -- & -- \\
        PointNet++~\cite{pointnet++} &  Point cloud & 89.7 & \textbf{94.6} & \textbf{76.4} & -- & -- \\
        DGCNN~\cite{dgcnn} &  Point cloud & \textbf{89.9} & 94.3 & 76.2 & -- & -- \\
        MeshWalker~\cite{meshwalker} &  Mesh &  -- & -- & -- & \textbf{90.0} & -- \\
        Conv3DNet~\cite{maturana2015voxnet}&  Voxels & -- & -- & -- & -- & 92.1 \\
        \cmidrule(lr){1-7}
        \inrtovec{} &  \nf{} & 87.0 & 93.3 & 72.1 & 86.3 & \textbf{93.0}\\
        \bottomrule
    \end{tabular}
    \caption{\textbf{Results on shape classification across representations.}}
    \label{tab:classification}
\end{table*}


\begin{table*}[t]
    \centering
    \resizebox{\linewidth}{!}{
    \begin{tabular}{llcccccccccccccccccc}
    Method & Input & \rotatebox{90}{instance mIoU} & \rotatebox{90}{class mIoU} & \rotatebox{90}{airplane} & \rotatebox{90}{bag} & \rotatebox{90}{cap} & \rotatebox{90}{car} & \rotatebox{90}{chair} & \rotatebox{90}{earphone} & \rotatebox{90}{guitar} & \rotatebox{90}{knife} & \rotatebox{90}{lamp} & \rotatebox{90}{laptop} & \rotatebox{90}{motor} & \rotatebox{90}{mug} & \rotatebox{90}{pistol} & \rotatebox{90}{rocket} & \rotatebox{90}{skateboard} & \rotatebox{90}{table}\\
    \midrule
    PointNet~\cite{pointnet} & Point cloud & 83.1 & 78.96 & 81.3 & 76.9 & 79.6 & 71.4 & 89.4 & 67.0 & 91.2 & 80.5 & 80.0 & 95.1 & 66.3 & 91.3 & 80.6 & 57.8 & 73.6 & 81.5\\
    PointNet++~\cite{pointnet++} & Point cloud & \textbf{84.9} & \textbf{82.73} & \textbf{82.2} & \textbf{88.8} & \textbf{84.0} & \textbf{76.0} & \textbf{90.4} & 80.6 & \textbf{91.8} & 84.9 & \textbf{84.4} & 94.9 & \textbf{72.2} & \textbf{94.7} & \textbf{81.3} & 61.1 & \textbf{74.1} & \textbf{82.3}\\ 
    DGCNN~\cite{dgcnn} & Point cloud & 83.6 & 80.86 & 80.7 & 84.3 & 82.8 & 74.8 & 89.0 & \textbf{81.2} & 90.1 & \textbf{86.4} & 84.0 & \textbf{95.4} & 59.3 & 92.8 & 77.8 & \textbf{62.5} & 71.6 & 81.1\\
    \cmidrule(lr){1-20}
    \inrtovec{} & \nf{} & 81.3 & 76.91 & 80.2 & 76.2 & 70.3 & 70.1 & 88.0 & 65.0 & 90.6 & 82.1 & 77.4 & 94.4 & 61.4 & 92.7 & 79.0 & 56.2 & 68.6 & 78.5\\
    \bottomrule
    \end{tabular}}
    \caption{\textbf{Part segmentation quantitative results.} We report the IoU for each class, the mean IoU over all the classes (class mIoU) and the mean IoU over all the instances (instance mIoU).}
    \label{tab:partseg_quantitatives}
\end{table*}

\begin{figure}[t]
    \centering
    \includegraphics[width=\linewidth]{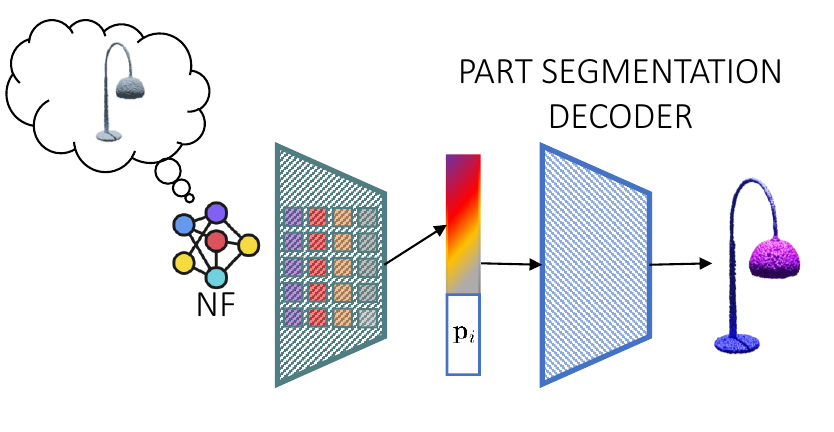}
    \caption{\textbf{Point cloud part segmentation.} Method.}
    \label{fig:partseg_method}
\end{figure}

\begin{figure}[t]
    \centering
    \includegraphics[width=0.9\linewidth]{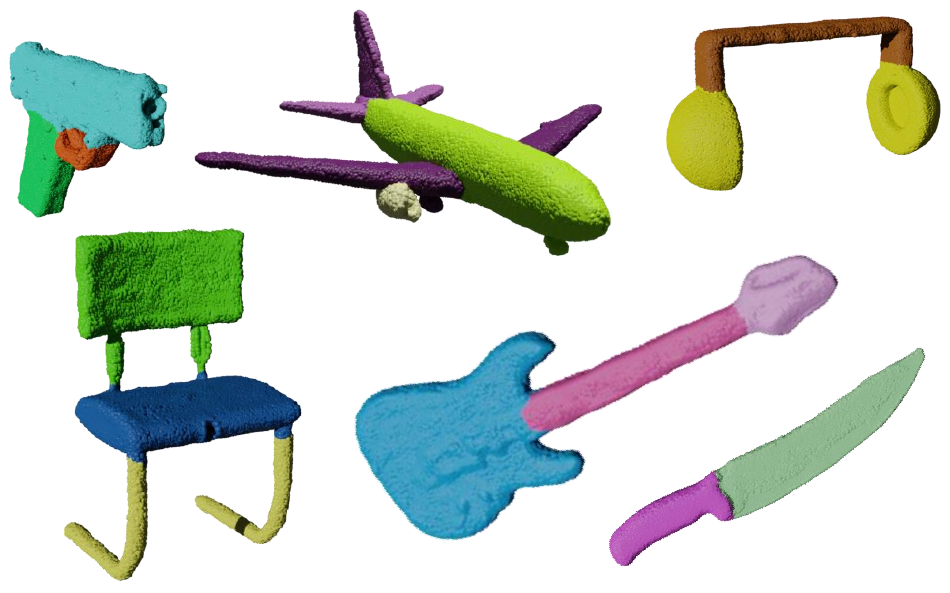}
    \caption{\textbf{Point cloud part segmentation.} Qualitatives.}
    \label{fig:partseg_qualitatives}
\end{figure}


\begin{figure*}
    \centering
    \begin{minipage}{0.3\linewidth}
    \includegraphics[width=\linewidth]{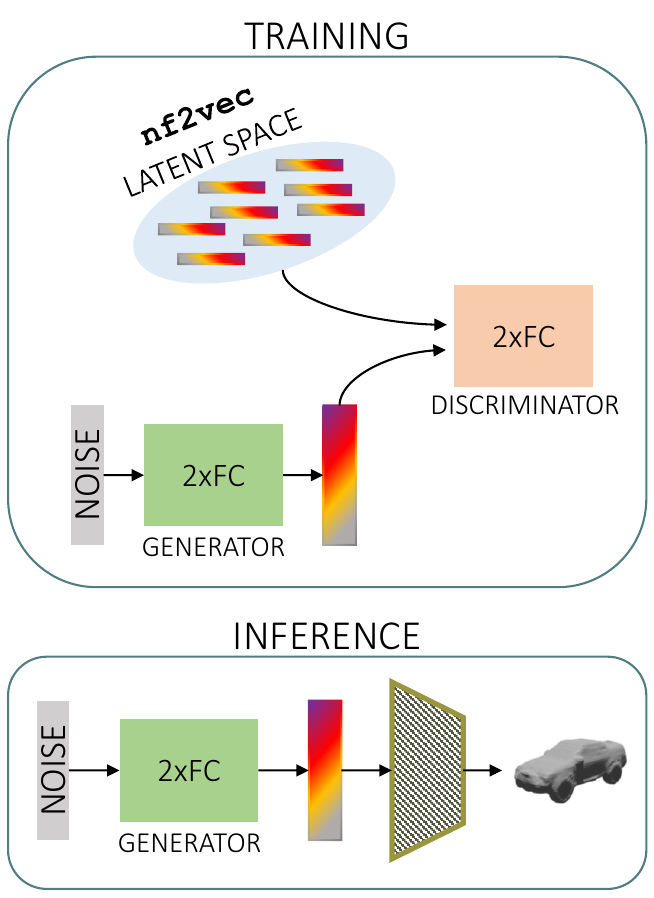}
    \caption{\textbf{Learning to generate shapes from \inrtovec{} latent space.} Method.}
    \label{fig:generative_method}
    \end{minipage}
    \hspace{1em}
    \begin{minipage}{0.6\linewidth}
        \centering
        \includegraphics[width=\linewidth]{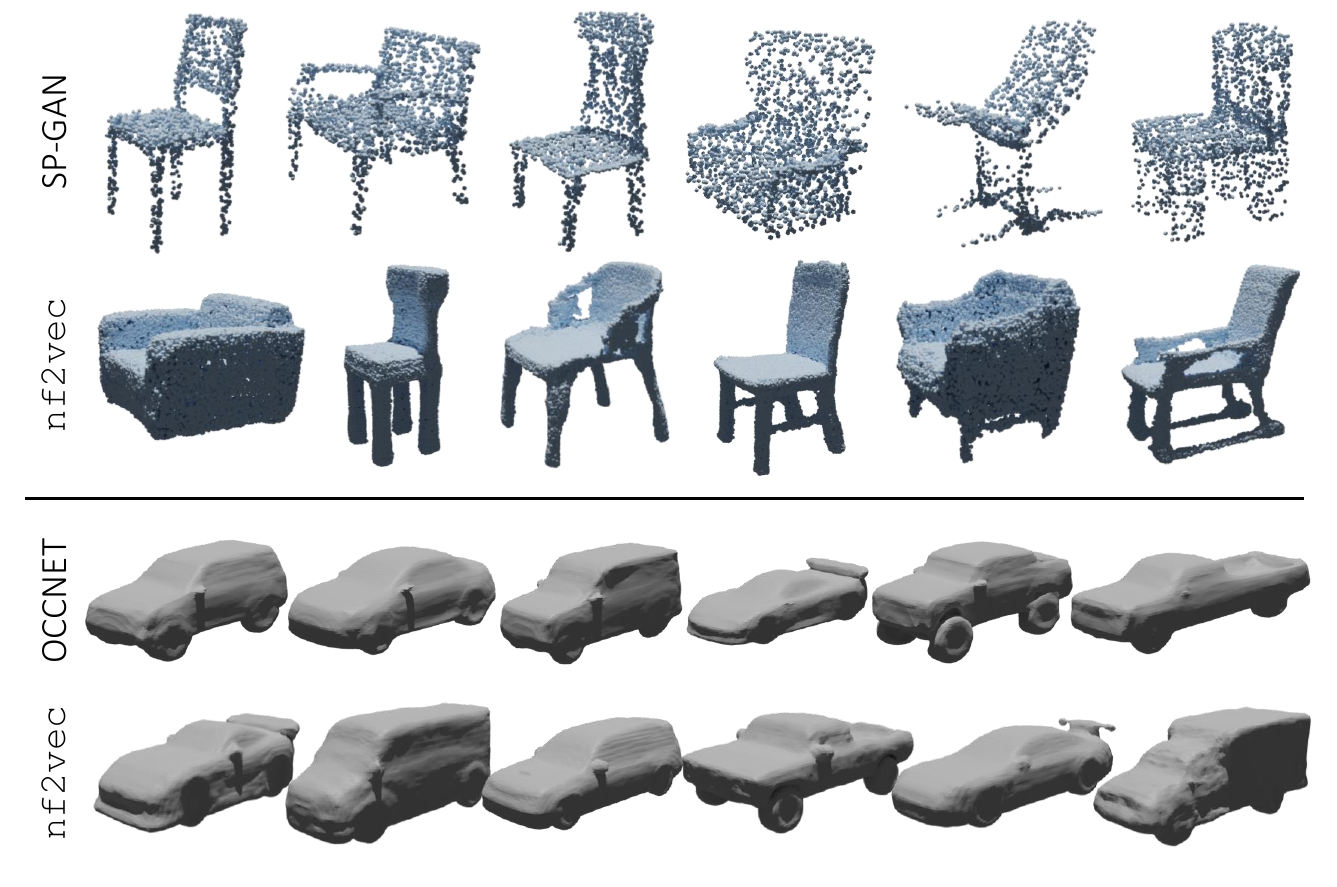}
        \caption{\textbf{Learning to generate shapes from \inrtovec{} latent space.} Qualitative results.}
        \label{fig:generative_qualitatives}
    \end{minipage}
\end{figure*}

\textbf{Shape classification.}
We then address the problem of classifying point clouds, meshes, and voxel grids.
We use three datasets for point clouds: ShapeNet10, ModelNet40, and ScanNet10 \cite{dai2017scannet}. When dealing with meshes, we conduct our experiments on the Manifold40 dataset \cite{subdivnet}. Finally, we use ShapeNet10 again for voxel grids, quantizing clouds to grids with resolution $64^3$.
Despite the different nature of the discrete representations taken into account, \inrtovec{} allows us to perform shape classification on \nf{}s embeddings, augmented online with E-Stitchup \cite{estitchup}, by the very same downstream network architecture, \ie{} a simple fully connected classifier consisting of three layers with 1024, 512 and 128 features.
We consider as baselines well-known architectures that are optimized to work on the specific input representations of each dataset.
For \pc{}, we consider PointNet \cite{pointnet}, PointNet++ \cite{pointnet++} and DGCNN \cite{dgcnn}. 
For meshes, we consider MeshWalker \cite{meshwalker}, a recent and competitive baseline that processes triangle meshes directly. As for voxel grids, we train a 3D CNN classifier that we implemented following  \cite{maturana2015voxnet} (Conv3DNet from now on).
Since only the train and test splits are released for all the datasets, we created validation splits from the training sets in order to follow a proper train/val protocol for both the baselines and our method. 
As for the test shapes, we evaluated all the baselines on the discrete representations reconstructed from the \nf{}s fitted on the original test sets, as these would be the only data available at test time in a scenario where \nf{}s are used to store and communicate 3D data.
%
The results in \cref{tab:classification} show that \inrtovec{} embeddings deliver classification accuracy close to the specialized baselines across all the considered datasets, regardless of the original discrete representation of the shapes in each dataset. Remarkably, our framework allows us to apply the same simple classification architecture to all the considered input modalities, in stark contrast with all the baselines that are highly specialized for each modality, exploit inductive biases specific to each such modality and cannot be deployed on representations different from those they were designed for. 
Furthermore, while presenting a gap of some accuracy points \wrt{} the most recent architectures, like DGCNN and MeshWalker, the simple fully connected classifier that we applied on \inrtovec{} embeddings obtains scores comparable to standard baselines like PointNet and Conv3DNet.

Finally, in \cref{fig:cls_inference_time_details} (left), we present the baseline inference times, assuming discrete point clouds are available at test time, and compare it with that of \inrtovec{}.
Our framework is much faster than competitors. Indeed, by processing directly \nf{}s -- where the resolution of the underlying signal is theoretically infinite -- \inrtovec{} can classify \nf{}s representing point clouds with different numbers of points with a constant inference time of 0.001 seconds. On the contrary, the examined baselines suffer from the escalating resolution of the input point clouds. While PointNet and PointNet++ maintain a reasonable inference time even with 64K points, DGCNN experiences a significant slowdown as early as 16K points. 
Additionally, we emphasize that if 3D shapes are stored as \nf{}s, the classification process using the designated specialized baselines would involve retrieving the original discrete representations through the extensive procedures outlined in \cite{deluigi2023inr2vec}.
Therefore, in \cref{fig:cls_inference_time_details} (right), we present the inference time of standard point cloud classification networks, including the time needed to reconstruct the discrete point cloud from the input \nf{} of the underlying $udf$ at various resolutions. Even at the coarsest resolution (2048 points), all the baselines exhibit an inference time that is one order of magnitude higher than the time required to classify the \inrtovec{} embeddings directly.
As the resolution of the reconstructed clouds increases, the inference time of the baselines becomes prohibitively high, while \inrtovec{}, not reliant on explicit clouds, maintains a constant inference time of 0.001 seconds.

\textbf{Point cloud part segmentation.}
The classification and retrieval tasks explore the potential of utilizing \inrtovec{} embeddings as a global representation of the input shapes. In contrast, in this section, we focus on point cloud part segmentation to examine if \inrtovec{} embeddings also retain local shape properties.
Part segmentation aims to predict a semantic (\ie{} part) label for each point of a given cloud. We tackle this problem by training a decoder similar to that used to train our framework (see \cref{fig:partseg_method}). Such decoder is fed with the \inrtovec{} embedding of the \nf{} representing the input cloud, concatenated with the coordinate of a 3D query, and it is trained to predict the label of the query point.
We train it, as well as PointNet, PointNet++, and DGCNN, on the ShapeNet Part Segmentation dataset \cite{shapenet-part} with point clouds of 2048 points, with the same train/val/test as in the classification task.
The outcomes presented in \cref{tab:partseg_quantitatives} demonstrate the potential of accomplishing a local discriminative task, such as part segmentation, by using the task-agnostic embeddings generated by \inrtovec{}. In doing so, the performance achieved is notably close to that of dedicated architectures designed for this specific task.
Additionally, in \cref{fig:partseg_qualitatives}, we show point clouds reconstructed at 100K points from the input \nf{}s and segmented with high precision thanks to our formulation based on a semantic decoder conditioned by the \inrtovec{} embedding.

\begin{figure*}
    \centering
    \begin{minipage}{0.28\linewidth}
            \includegraphics[width=\textwidth]{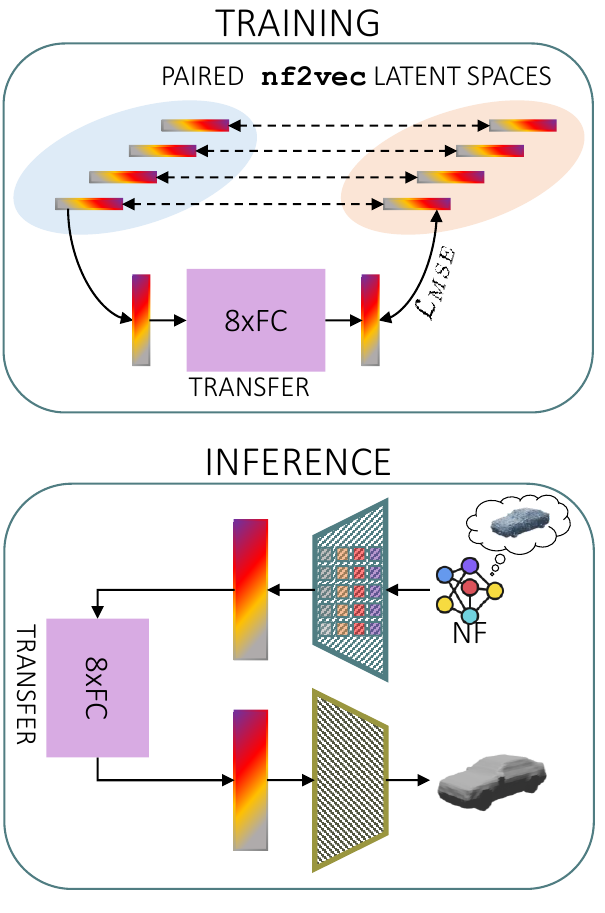}
            \caption{\textbf{Learning a mapping between \inrtovec{} latent spaces.} Method.}
            \label{fig:mapping_method}
    \end{minipage}
    \hspace{1em}
    \begin{minipage}{0.6\linewidth}
            \includegraphics[width=\textwidth]{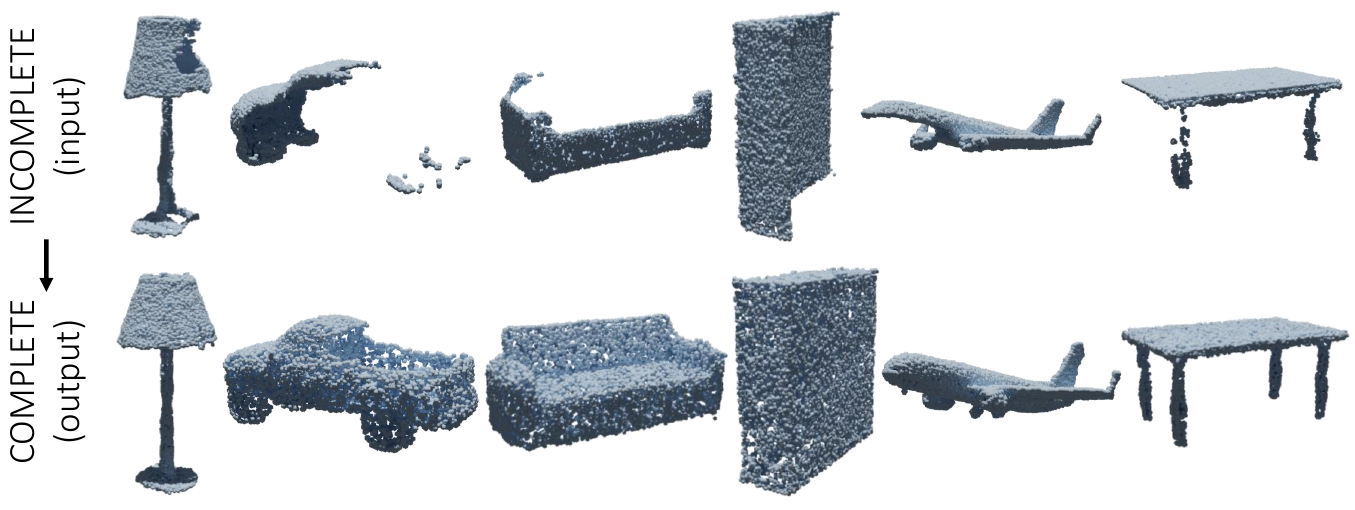}
            \caption{\textbf{Learning a mapping between \inrtovec{} latent spaces.} Point cloud completion.}
            \label{fig:completion_qualitatives}
            \includegraphics[width=\textwidth]{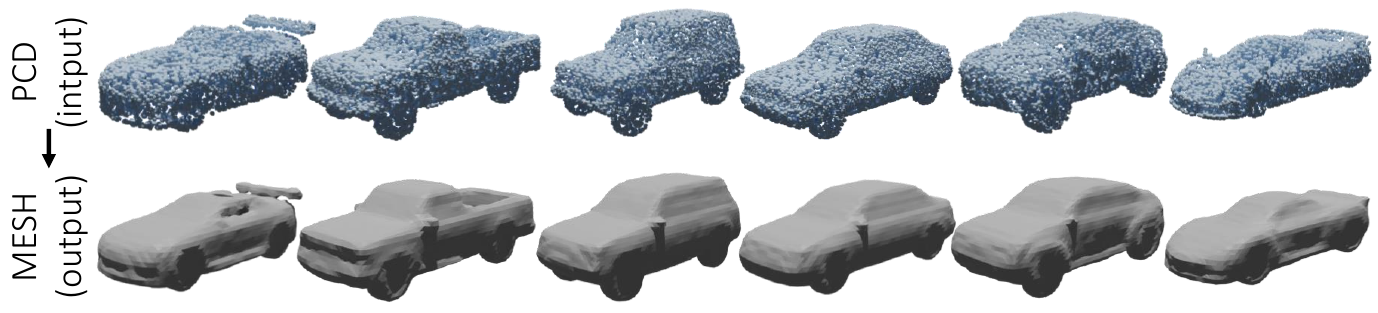}
            \caption{\textbf{Learning a mapping between \inrtovec{} latent spaces.} Surface reconstruction.}
            \label{fig:surfrec_qualitatives}
    \end{minipage}
\end{figure*}

\textbf{Shape generation.}
So far, we have validated that \nf{} can be used as input in standard deep learning machinery thanks to \inrtovec{}. In this section, we focus on the task of shape generation in an adversarial setting to investigate whether the compact representations produced by our framework can also be adopted as a medium for the output of generative deep learning pipelines.
We employ a Latent-GAN \cite{latent-gan} to generate embeddings resembling those produced by \inrtovec{} from random noise, as illustrated in \cref{fig:generative_method}. Generated embeddings can be decoded into discrete representations using the implicit decoder from the \inrtovec{} training.
As our framework is agnostic towards the original discrete representation of shapes used to learn the \nf{}s, we can train Latent-GANs with embeddings representing point clouds or meshes based on the same identical protocol and architecture (two simple fully connected networks as generator and discriminator).
For point clouds, we train a Latent-GAN on the \textit{chair} class of ShapeNet10, while we use models of cars provided by \cite{occupancynetworks} when dealing with meshes.
In \cref{fig:generative_qualitatives}, we report some examples generated with the above procedure, comparing them with SP-GAN~\cite{spgan} for what concerns point clouds and Occupancy Networks~\cite{occupancynetworks} (VAE formulation) for meshes. 
Shapes generated by our Latent-GAN, trained exclusively on \inrtovec{} embeddings, look quite similar to those from the considered baselines regarding diversity and richness of details. Furthermore, by generating embeddings representing \nf{}s,  our method allows point cloud sampling at any arbitrary resolution (\eg{} 8192 points in \cref{fig:generative_qualitatives}). In contrast, SP-GAN needs a new training for each desired resolution, as the number of generated points must be predetermined during training.

\textbf{Learning a mapping between \inrtovec{} embedding spaces.}
We have shown that \inrtovec{} embeddings can be employed as a proxy of \nf{}s as input to deep learning pipelines and that they can also be obtained as output of generative frameworks.
In this section, we advance our exploration by examining the potential of learning a mapping between two distinct latent spaces generated by our framework for two separate datasets of \nf{}s. This involves developing a \textit{transfer} function specifically designed to operate on \inrtovec{} embeddings as both input and output data.
Such transfer function can be realized by a simple MLP that maps the input embedding into the output one and is trained with standard MSE loss (see \cref{fig:mapping_method}).
As \inrtovec{} generates compact embeddings of the same dimension regardless of the input \nf{} modality, the transfer function described here can be applied seamlessly to a great variety of tasks, usually tackled with ad-hoc frameworks tailored to specific input/output modalities.
In particular, We explore two tasks. First, on the dataset presented in \cite{vrcnet}, we address point cloud completion by learning a mapping from \inrtovec{} embeddings of \nf{}s that represent incomplete clouds to embeddings associated with complete clouds. Then, we tackle the task of surface reconstruction on ShapeNet cars, training the transfer function to map \inrtovec{} embeddings representing point clouds into embeddings that can be decoded into meshes.
As we note from \cref{fig:completion_qualitatives} and \cref{fig:surfrec_qualitatives}, in both tasks, the transfer function can learn an effective mapping between \inrtovec{} latent spaces. Indeed, by processing exclusively \nf{} embeddings, we can obtain output shapes that are highly compatible with the input ones whilst preserving their distinctive details, like the pointy wing of the airplane in \cref{fig:completion_qualitatives} or the flap of the first car in \cref{fig:surfrec_qualitatives}.

\section{Deep Learning on NeRFs}
\label{sec:experiments_nerf}

In this section, our focus shifts to processing \nf{}s encoding both geometry and appearance of objects, i.e., NeRFs. The goal is to illustrate the efficacy of \nerfalgoname{} in addressing various downstream tasks related to 3D objects implicitly represented by NeRFs.

\begin{table}
    \centering
    \resizebox{\linewidth}{!}{
    \begin{tabular}{lcccc}
        \toprule
        Method & \#Views & mAP@1 & mAP@5 & mAP@10 \\
        \midrule
        \nerfalgoname{} & - & 72.38 & \textbf{91.89} & \textbf{95.96} \\
        \cmidrule(lr){1-5}
        ResNet50 \cite{resnet} single-view & 1 & 74.65 & 91.52 & 95.10 \\
        ResNet50 \cite{resnet} multi-view & 9 & \textbf{82.74} & 91.66 & 93.79 \\
        \bottomrule
    \end{tabular}}
    \caption{\textbf{NeRF retrieval quantitative results.}}
    \label{tab:nerf_retrieval}
\end{table}
\begin{table}[t]
    \centering
    \begin{tabular}{lcc}
        \toprule
        Method & \#Views & Accuracy (\%)\\
         \midrule
         \nerfalgoname{} & - & 87.28 \\
         \cmidrule(lr){1-3}
         ResNet50 \cite{resnet} Single-view & 1  & 86.88 \\
         ResNet50 \cite{resnet} Multi-view & 9 & \textbf{93.28} \\
         \bottomrule
    \end{tabular}
    \caption{\textbf{Results on NeRF classification}.}
    \label{tab:nerf_classifier}
\end{table}
\textbf{General settings.} In all experiments detailed within this section, we learn NeRFs from images using an MLP comprising three hidden layers with 64 nodes each. We utilize the ReLU activation function between all layers except the final layer, which computes the density and RGB values without any activation function. NeRFs take as input the frequency encoding of the 3D coordinates as in \cite{nerf}. NeRFs are trained using an $L_1$ loss between predicted and estimated RGB pixel intensities, weighting background pixels less than foreground pixels (0.8 foreground vs 0.2 background).
We use the NeRF formulation without the view direction in input.
When training \nerfalgoname{}, we adhere to the same encoder and decoder architectures previously described in \cref{sec:experiments}. The key distinction lies in the implicit decoder, wherein the dimensions of its layers are doubled. Throughout these experiments, both baseline models and \nerfalgoname{} undergo training with offline data augmentation on the 3D shapes used to generate renderings for training the models. This augmentation involves implementing random deformations and introducing random color changes to each component of the original 3D shapes.

\begin{table}
    \centering
    \resizebox{\linewidth}{!}{
    \begin{tabular}{lcccc}
    \toprule
        & \multicolumn{4}{c}{Time (ms)} \\
        \cmidrule(lr){2-5}
        Method & Encoding & Rendering & Classification & Total\\
        \midrule
        \nerfalgoname{} & 0.75 & - & 0.28 & \textbf{1.03} \\
        \cmidrule(lr){1-5}
        Resnet50 \cite{resnet} Single-View & - & 4.63 & 6.21 & 10.84 \\
        Resnet50 \cite{resnet} Multi-View & - & 41.67 & 55.89 & 97.56 \\
        \bottomrule
    \end{tabular}}
    \caption{\textbf{NeRFs Classification Inference Time.} All times are in milliseconds. For each method, we report the time needed for each pipeline step and the total time. Encoding: \nerfalgoname{} encoding of \nf{} weights. Rendering: obtaining images from NeRFs. Classification: the time required by the classifier.}
    \label{tab:nerf_classifier_timings}
\end{table}
\begin{figure}
    \centering
    \includegraphics[width=1.0\linewidth]{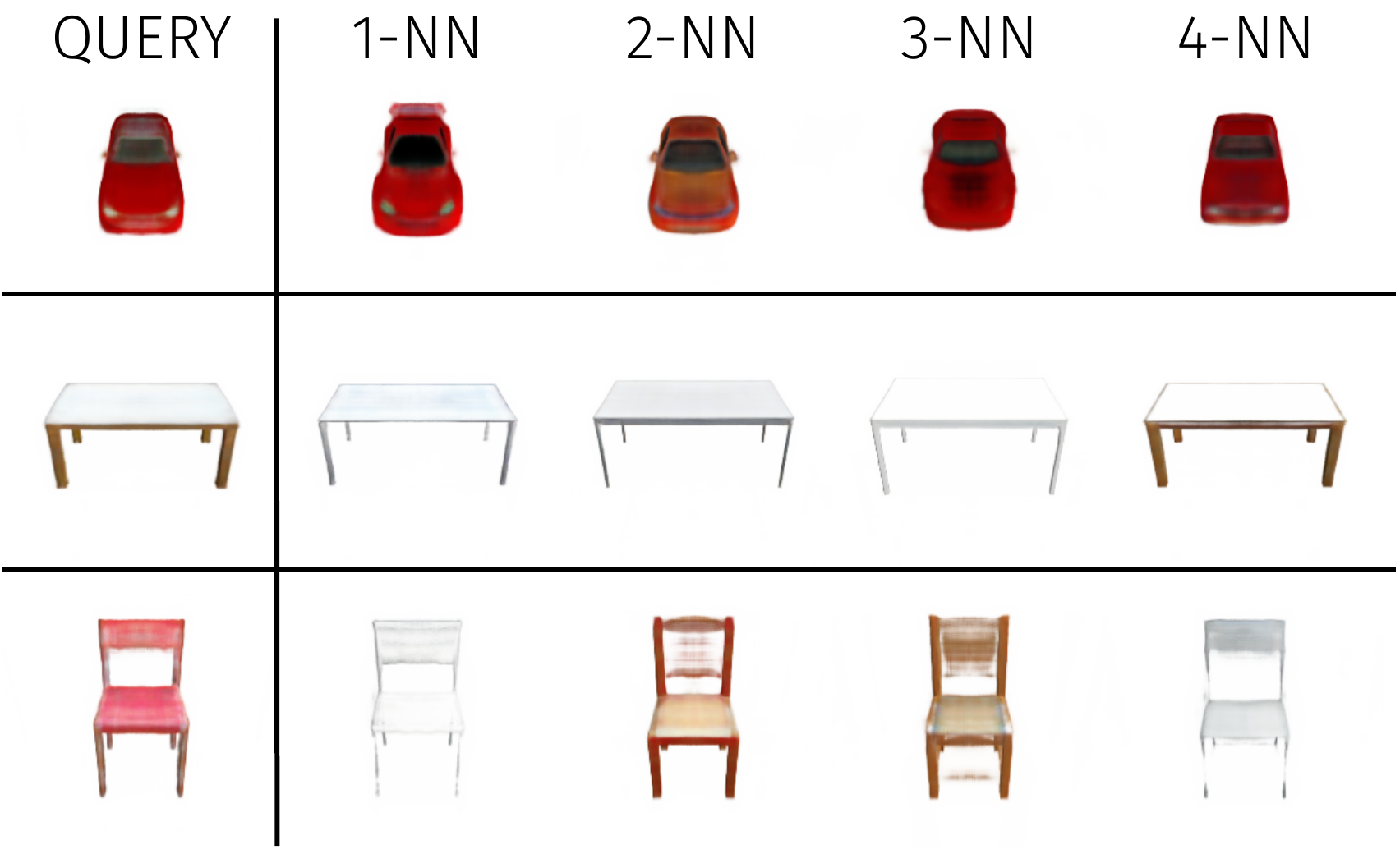}
    \caption[NeRF retrieval qualitative results]{\textbf{NeRF retrieval qualitative results.} Given the \nerfalgoname{} embedding of a query NeRF, we show the renderings reconstructed from the closest embeddings (L2 distance).}
    \label{fig:nerf_retrieval}
\end{figure}

\textbf{NeRF retrieval.} We first investigate the quality of \nerfalgoname{} embeddings with a retrieval task as done in \cref{sec:experiments}.
Given an embedding of an input NeRF, we extract the classes of its \textit{k}-nearest neighbor within the embedding latent space and compare them to the input NeRF’s class.

We also implement two baseline approaches: the single-view and multi-view baselines. Both strategies rely on a \textit{ResNet50} \cite{resnet} backbone pre-trained on ImageNet \cite{imagenet}. We extract feature vectors with \textit{ResNet50} from each image. Given a single image for the single-view baseline or 9 images for the multi-view baseline, we find the \textit{k}-nearest neighbors in the \textit{ResNet50} feature space. We compare the classes of query and retrieved objects and compute the mAP for different $k$ as done in \cref{tab:retrieval}. In the case of the multi-view baseline, we retrieve the nearest neighbors for each of the 9 input images. Then, we select the class with the highest frequency to calculate the mAP.
All these experiments are conducted on embeddings of unseen NeRFs from the test set of ShapeNetRender \cite{shapenet_render}.
The quantitative results in Table \ref{tab:nerf_retrieval} indicate that \nerfalgoname{} yields comparable performance to the baselines and, in certain instances,  outperforms them. Moreover, Fig. \ref{fig:nerf_retrieval} shows that the selected neighbors exhibit similar structures and colors. There is, however, no way to prioritize one property over the other, a limitation discussed in \cref{sec:limitations}.

\textbf{NeRF classification.} 
This section investigates the task of predicting the category of an object represented by a NeRF. 
In this scenario, only NeRFs would be available as input data.

Our approach processes \nerfalgoname{} embeddings with the same classification architecture as already deployed for shapes, a three-layer MLP with 1024, 512, and 128 neurons each.
We highlight that the embeddings are obtained by processing the NeRFs weights without sampling the underlying \nerf{}.

As the discrete representations used to learn NeRFs are a set of images depicting the same object, selecting a proper baseline is not straightforward. In our experiment, we choose ResNet50 \cite{resnet} as the baseline classifier. The network predicts the class for a given input image. Given this architecture, akin to the retrieval experiment, we propose two types of baseline approaches, single-view and multi-view.
In the former, we train the network on a single rendering for each NeRF obtained from the same fixed pose, while for the latter, we employ 9 renderings for each NeRF from different viewpoints. 
At test time, regarding the single-view approach, we test the network on images rendered from unseen NeRFs employing the same training pose. Concerning the multi-view baseline, we render 9 images from the training viewpoints for each unseen NeRF, obtaining 9 distinct predictions per object, that we aggregate by taking the class predicted with the highest frequency.

We report the accuracy results on ShapeNetRender \cite{shapenet_render} in \cref{tab:nerf_classifier}. Moreover, we also report the time required to classify NeRFs in \cref{tab:nerf_classifier_timings}, highlighting the impact of each of the main pipeline steps, such as the \nerfalgoname{} encoding time for our approach and the time to render images for the baseline approaches. We point out that we rely on a fast implementation of NeRFs provided by the NerfAcc \cite{nerfacc} library.
We note that the classifier directly leveraging \textit{nerf2vec} embeddings achieves a slightly better score than the single-view baseline while being worse than the multi-view one.
However, if we analyze the total inference times, our approach is two orders of magnitude faster than the baselines (1.03ms our vs 97.56ms ResNet50 multi-view). This holds true even excluding the rendering time and looking only at the classification time (third column of \cref{tab:nerf_classifier_timings}): ResNet50 classifiers processing $224 \times 224$ images are remarkably slower than our full pipeline, taking only 1.03ms.

\begin{figure}[t]
    \centering
    \includegraphics[width=1.0\linewidth]{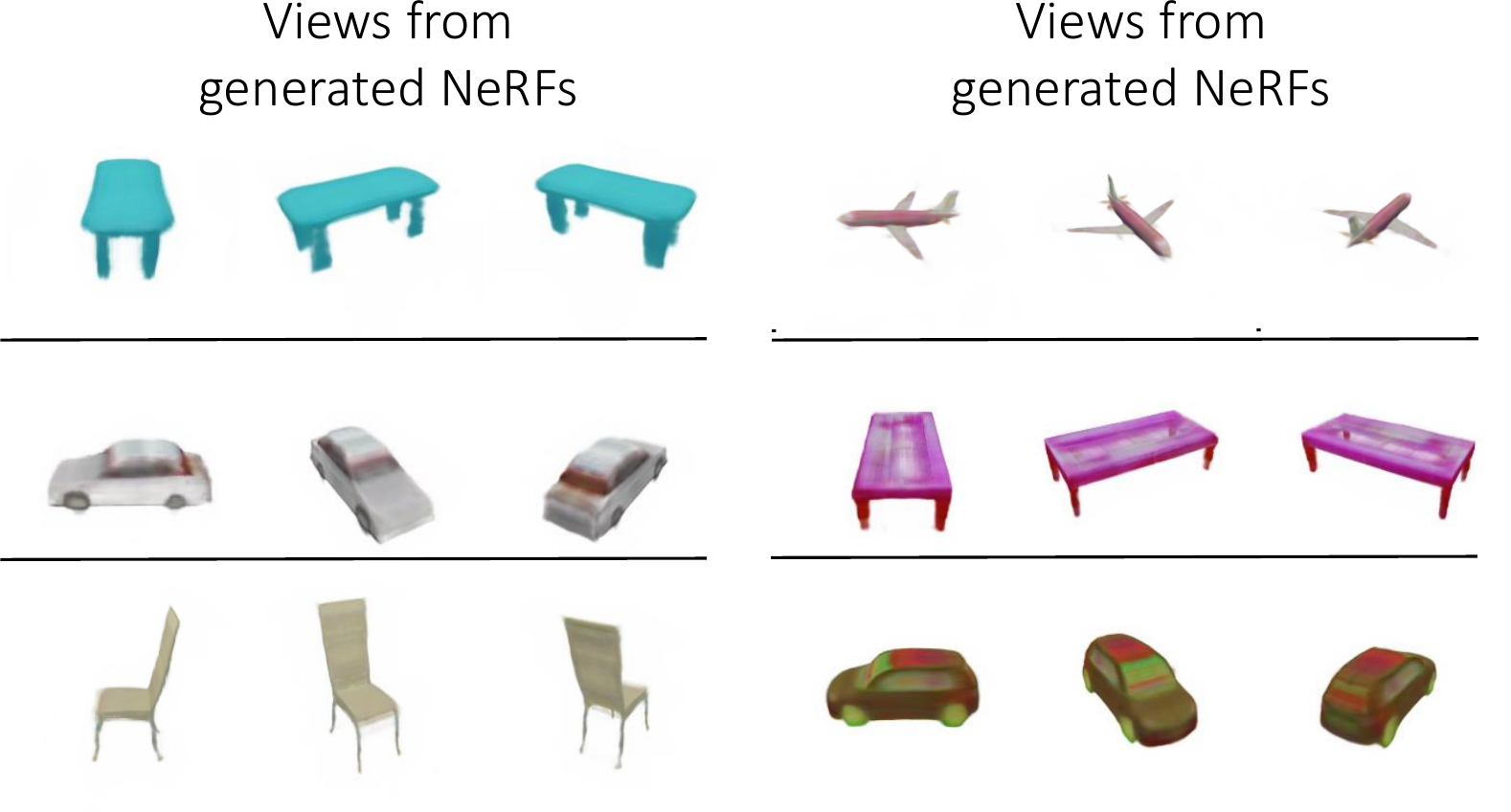}
    \caption{\textbf{Learning to generate NeRFs from \nerfalgoname{} latent space.} Qualitative results.}
    \label{fig:nerf_generations}
\end{figure}

\textbf{NeRF generation.} We experiment here with the task of generating \nerfalgoname{} embeddings with the same approach depicted in \cref{fig:generative_method}. We trained multiple adversarial networks, one for each class, utilizing the ShapeNetRender dataset \cite{shapenet_render}.
After the embedding generation, we can decode them into discrete representations using the same implicit decoder employed to train the framework.
In \cref{fig:nerf_generations}, we show images rendered from NeRFs generated through this process. 
The renderings have a good level of realism and diversity. Notably, the 3D consistency of images obtained from different viewpoints is preserved.
These results show that generating novel NeRFs in the form of \nerfalgoname{} embeddings is possible.

\begin{table}
    \centering
    \resizebox{\linewidth}{!}{
    \begin{tabular}{llcccc}
        \toprule
        &&&\multicolumn{2}{c}{Mesh Reconstruction} \\
        \cmidrule(lr){4-5}
        Method & Type & \#Params & CD (mm) $\downarrow$ & F-score (\%) $\uparrow$ \\
        \midrule
        DeepSDF \cite{deepsdf} & Shared MLP + Embedding & 2400K & 6.6 & 25.1 \\
        Functa \cite{functa} & Shared MLP + Modulation & 7091K & 2.85 & 21.3 \\
        \cmidrule(lr){1-5}
        DWS \cite{navon2023equivariant} & Individual SIREN MLP & \textbf{800K} & \textbf{0.26} & \textbf{69.7} \\
        \algoname{} & Individual SIREN MLP & \textbf{800K} & \textbf{0.26} & \textbf{69.7} \\
        \bottomrule
    \end{tabular}}
    \caption{\textbf{Properties of input \nf{}s used by recent methods processing neural fields.} Mesh reconstruction results on Manifold40 test set.}
    \label{tab:neural_functional_reconstruction}
\end{table}

\textbf{Learning a mapping between embedding spaces.}
We explore here whether it is possible to learn a \textit{transfer} network that maps \inrtovec{} embeddings of \udf{}s to \nerfalgoname{} embeddings of NeRFs using the methodology described in \cref{fig:mapping_method}. We highlight that it is a non-trivial task, as the network has to hallucinate the appearance of the input shape without modifying the underlying 3D structure.

\begin{figure}[t]
    \centering
    \includegraphics[width=0.8\linewidth]{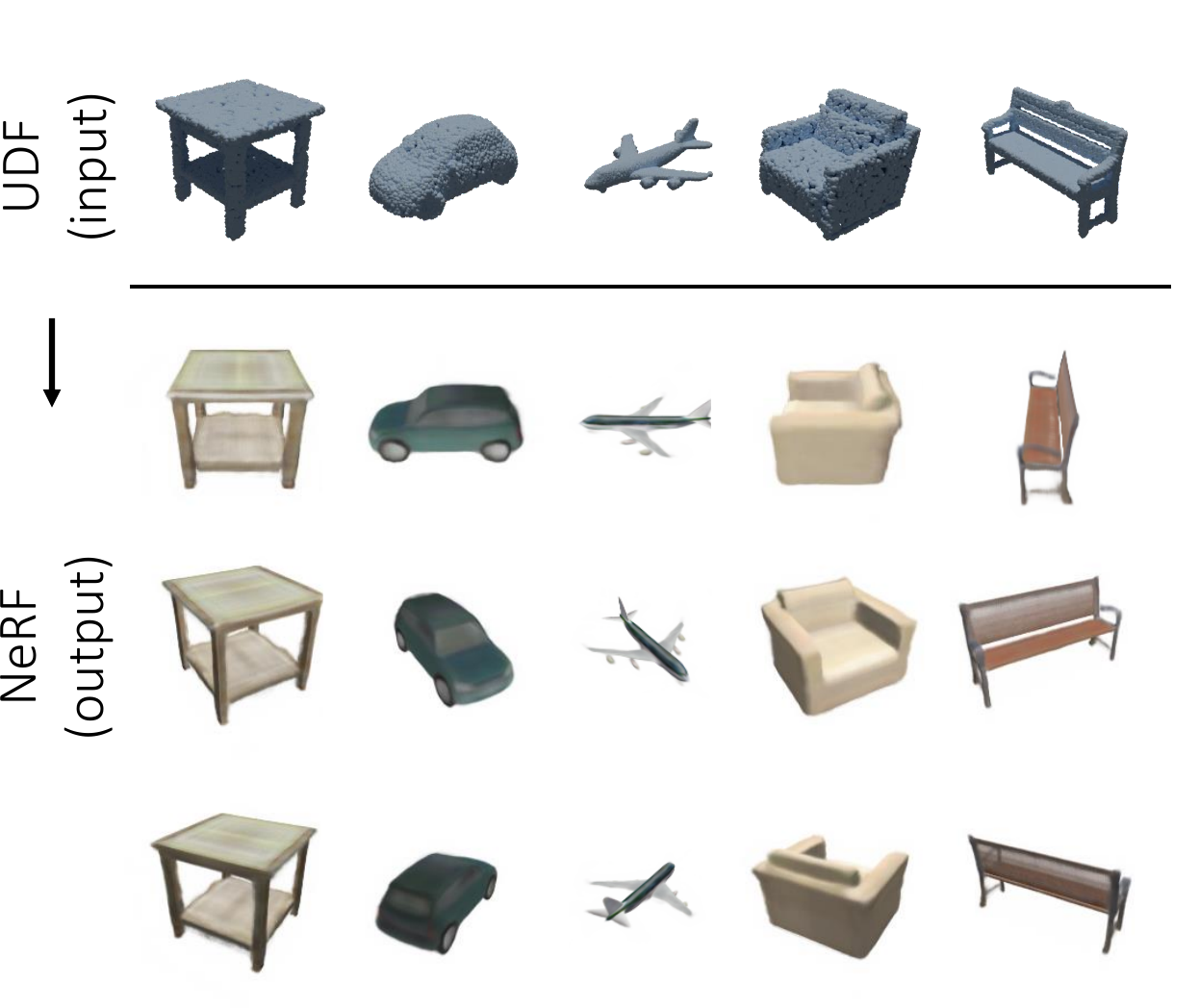}
    \caption{\textbf{Learning a mapping between \nerfalgoname{} latent spaces.} \udf{} to NeRF.}
    \label{fig:udf2nerf}
\end{figure}

We run this experiment on ShapeNetRender\cite{shapenet_render}, using the rendered images to learn NeRFs and the corresponding 3D models to learn \udf{} neural fields. Then, we train \nerfalgoname{} and then the \textit{transfer} network on \nf{}s of the training set.
Finally, we test the \textit{transfer} network on unseen test \nf{}, obtaining for each \udf{} latent code the corresponding mapped \nerfalgoname{} embedding. 
We report qualitative results in \cref{fig:udf2nerf}, showing the point clouds obtained from the input \udf{} in the first row, and the renderings obtained by feeding the mapped embedding to the implicit decoder of \nerfalgoname{} in the last three rows.
We can appreciate that the mapped NeRF preserves the original shape geometry, enabling the rendering of realistic images from various viewpoints. Notably, 
the renderings exhibit plausible diverse colors associated with different object parts, as can be seen for the glasses and the wheels of the blue car in the second column of \cref{fig:udf2nerf}.

\section{Comparison with Recent Approaches}
\label{sec:comparison_with_recent}
As outlined in \cref{sec:intro}, several contemporary works addressing the problem of processing neural fields have been proposed recently. For all methods, the goal is to perform deep learning tasks such as classification using as input data a \nf{}, \ie{} data represented with continuous functions.
We can divide these methods into two categories. Those relying on a shared network and those focusing on individual \nf{}s. In the former case, referred to here as  \textit{Shared}, the \nf{} is defined as a shared network trained on all training samples, plus a distinct vector representing each object. Typically this vector is processed to perform downstream tasks. This is the case of Functa \cite{functa} and DeepSDF \cite{deepsdf}. In the latter case, denoted as \textit{Individual}, the \nf{} is typically an MLP trained on a single object or scene. In this scenario, the MLP weights are processed directly to perform the downstream tasks. This is the case of our framework, \algoname{}, as well as NFN \cite{zhou2023permutation}, NFT \cite{zhou2023neural}, and DWS \cite{navon2023equivariant}.

In this section, we investigate the characteristics of each category of techniques, showing that \textit{Shared} frameworks are problematic, as they cannot reconstruct the underlying signal with high fidelity and need a whole dataset to learn the neural field of an object. Moreover, we build the first benchmark of \nf{} classification by comparing recent approaches in this area.

\begin{figure}
    \centering
    \includegraphics[width=\linewidth]{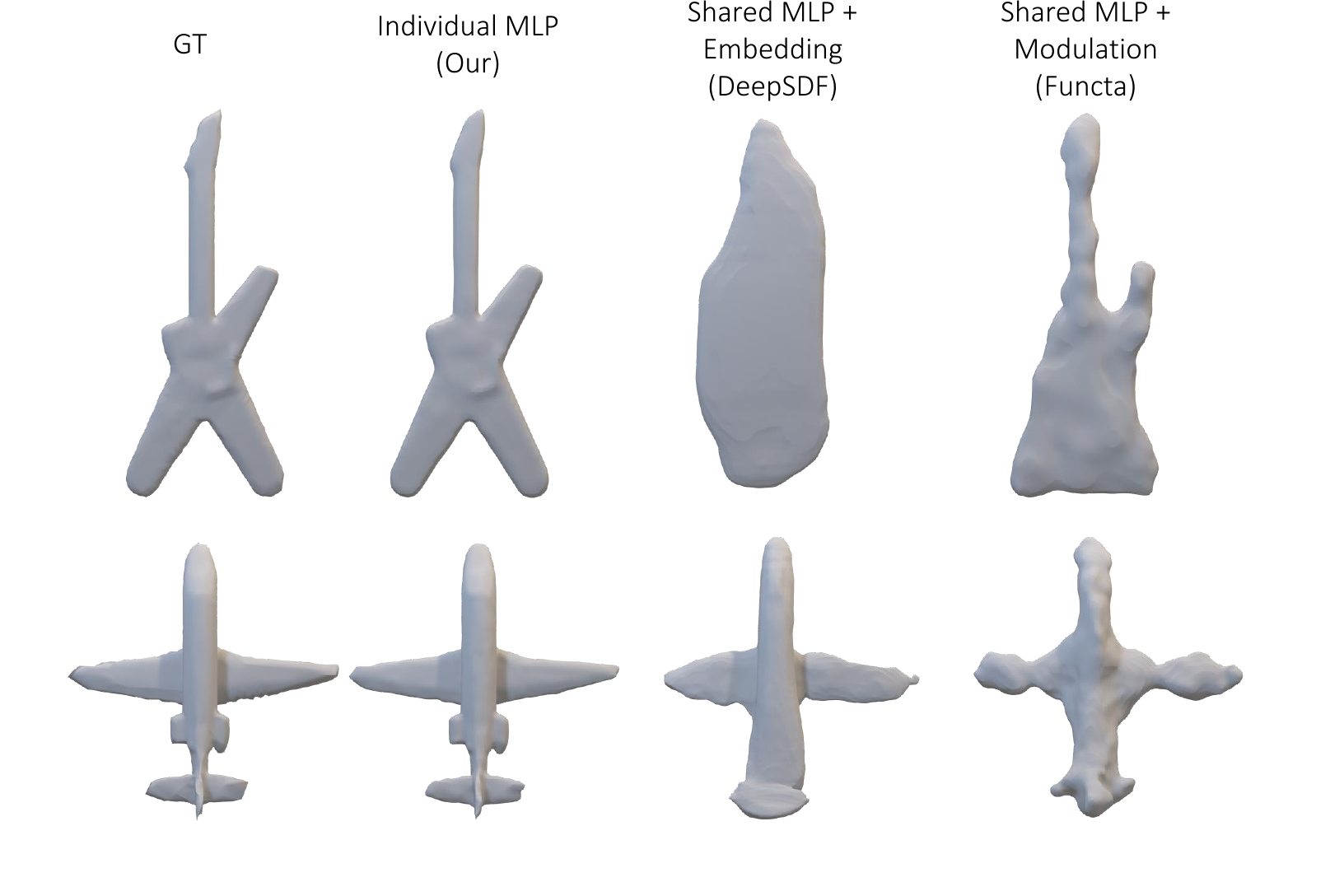}
    \caption{\textbf{Reconstruction comparison for Manifold40 meshes obtained from \sdf{}.}}
    \label{fig:mesh_rec}
\end{figure}

\begin{figure}
    \centering
    \includegraphics[width=\linewidth]{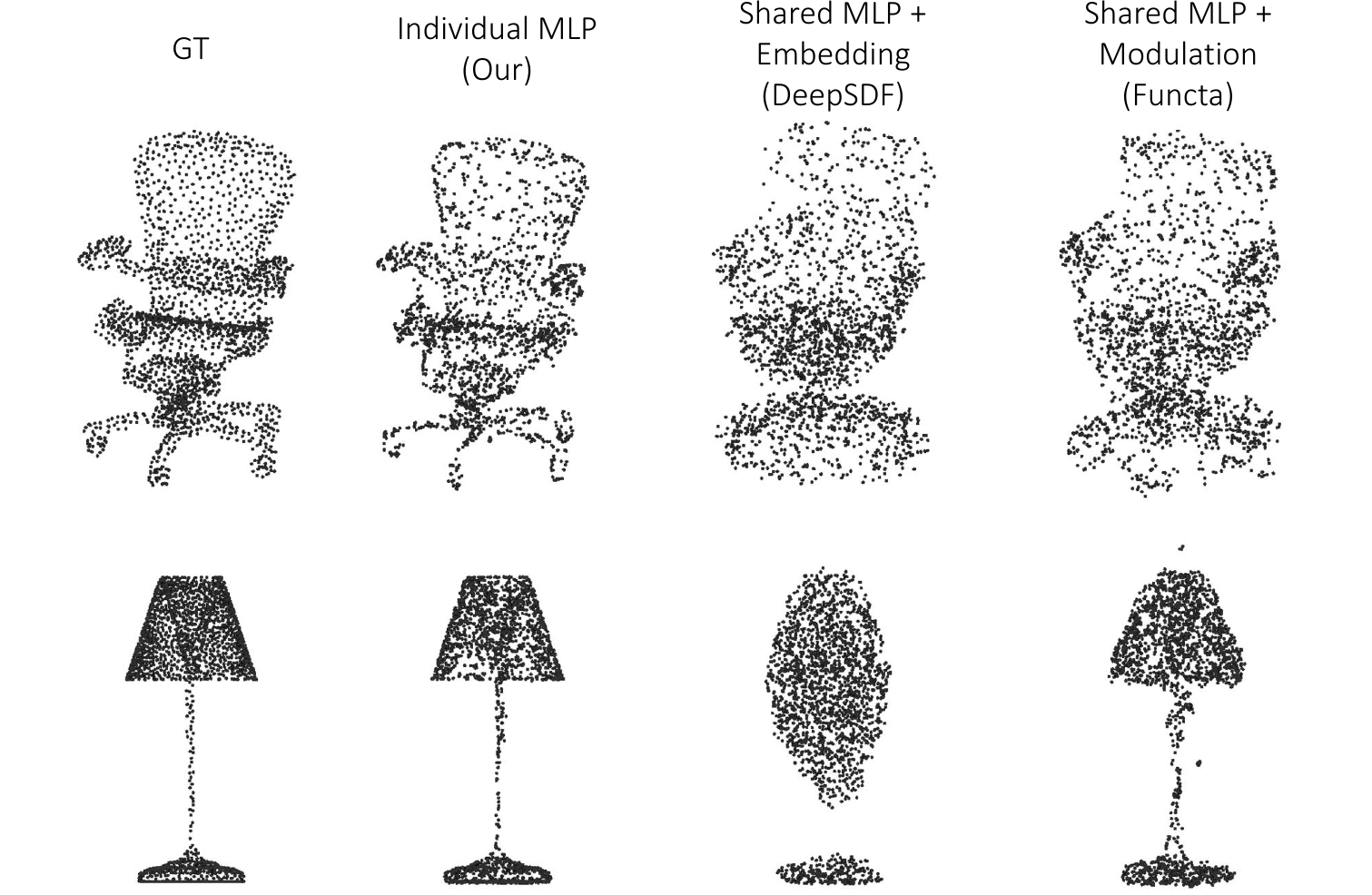}
    \caption{\textbf{Reconstruction comparison for ModelNet40 point clouds obtained from \udf{}.}}
    \label{fig:pcd_rec}
\end{figure}

\begin{table}[t]
    \centering
    \resizebox{\linewidth}{!}{
    \begin{tabular}{lllcc}
        \toprule
        Method & Type & Input & ModelNet40 (\udf{}) & Manifold40 (\sdf{})  \\
        \midrule
        DeepSDF \cite{deepsdf} & Shared & Embedding & 41.2 & 64.9\\
        Functa \cite{functa} & Shared & Modulation & \textbf{87.3} & 85.9 \\
        \cmidrule(lr){1-5}
        DWS \cite{navon2023equivariant} & Individual & MLP weights & 71.6 & 69.9 \\
        \algoname{} & Individual & MLP weights & 87.0 & \textbf{86.3} \\
        \bottomrule
    \end{tabular}}
    \caption{\textbf{Classification accuracy of recent methods processing neural fields.}}
    \label{tab:neural_functional_classification}
\end{table}
\textbf{Representation quality.}
We first investigate the representation quality of \textit{Shared} approaches compared to \textit{Individual} ones.
Specifically, we compare the reconstructions of explicit meshes from \sdf{} neural fields with ground truth meshes on the Manifold40 test set. We report the quantitative comparisons in \cref{tab:neural_functional_reconstruction}, using two metrics: the Chamfer Distance (CD) as defined in \cite{fan2017point}, and the F-Score as defined in \cite{tatarchenko2019single}. We note that we use the SIREN MLP described in \cref{sec:experiments} to represent \sdf{} with \textit{Individual} frameworks.
In the first two rows, we note that \textit{Shared} methods achieve poor reconstruction performance. Indeed, we believe that representing a whole dataset with a shared network is a difficult learning task, and the network struggles to fit accurately the totality of the samples. 
\textit{Individual} methods instead do not suffer from this problem and achieve very good reconstruction performance.
Moreover, we believe that the approaches based on \textit{Shared} networks struggle to represent unseen samples the further they are from the training distribution. Hence, in the foreseen scenario where \nf{}s become a standard representation for 3D data hosted in public repositories, leveraging on a single shared network may imply the need to frequently retrain the model upon uploading new samples, which, in turn, would change the embeddings of all the previously stored data. On the contrary, uploading the repository with a new object would not cause any sort of issue with individual \nf{}s, where one learns a network for each data point. 
Finally, we also provide a qualitative perspective of the aforementioned problem in \cref{fig:mesh_rec} and \cref{fig:pcd_rec}. The visualizations confirm the results of \cref{tab:neural_functional_reconstruction}, with shared network frameworks struggling to represent properly the ground-truth shapes, while individual \nf{}s enable high-fidelity reconstructions. We note that the quality of our DeepSDF reconstructions, where a single model is trained on the whole dataset, is inferior to the one reported in \cite{deepsdf}, where instead a different auto-decoder is trained for each class. This approach is not applicable in our case, as it would require to know in advance the class label of each shape in order to choose the right auto-decoder.

We believe that these results highlight that frameworks based on a single shared network cannot be used as a medium to represent objects as \nf{}s, because of their limited representation power when dealing with large and varied datasets and because of their difficulty in representing new shapes not available at training time.

\begin{figure}
    \centering
    \includegraphics[width=\linewidth]{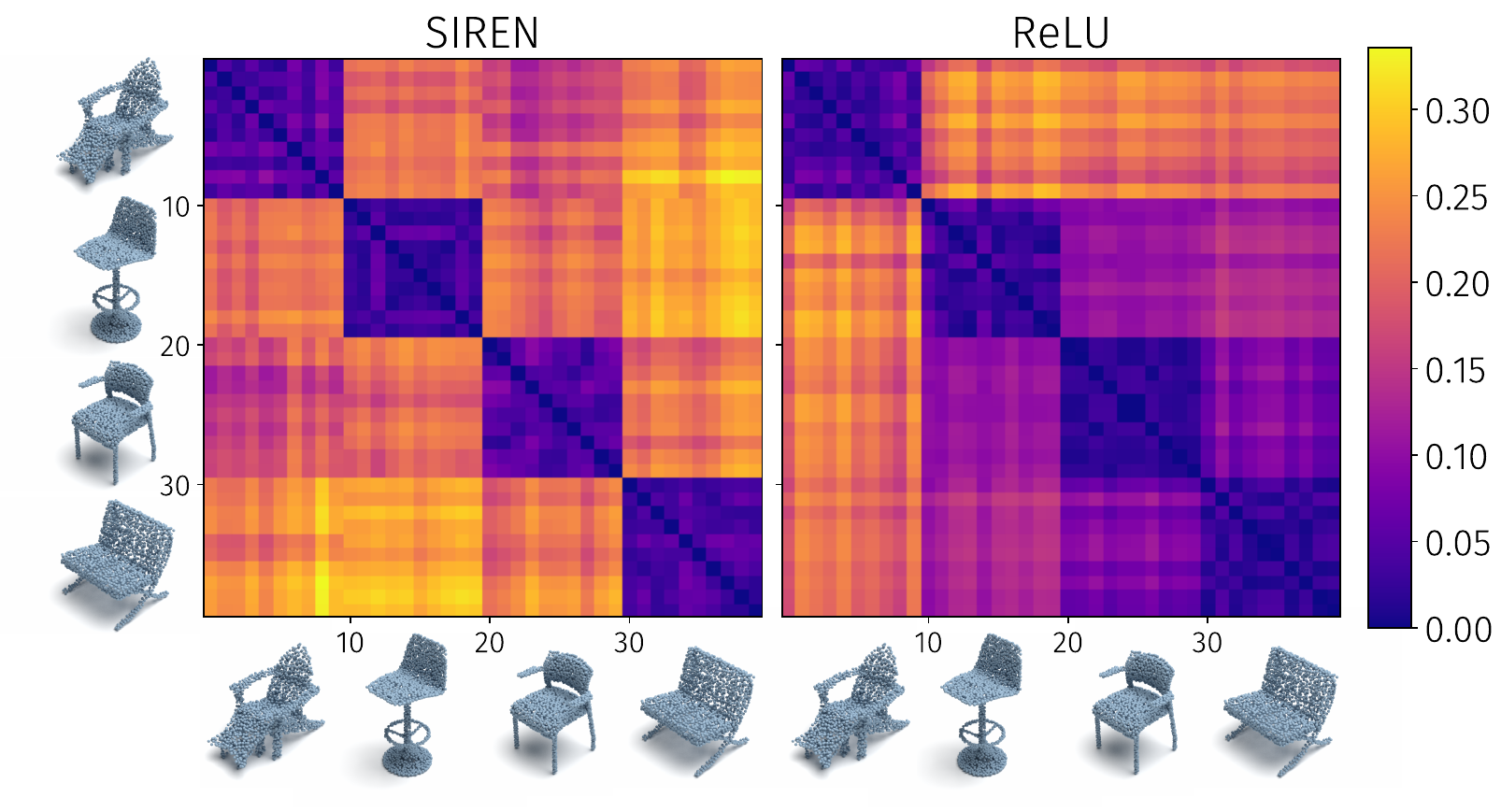}
    \caption{\textbf{L2 distances between \algoname{} embeddings.} For each shape, we fit 10 \nf{}s starting from the same weight initialization (40 \nf{}s in total). We then plot the L2 distances between the embeddings obtained by \algoname{} for such \nf{}s.}
    \label{fig:init_distances}
\end{figure}

\textbf{Classification accuracy.}
We compare recent methods in the \nf{}s classification task. The goal is to predict the category of objects represented within the input \nf{}s without recreating the discrete signals.
Specifically, we test all methods on \udf{} obtained from point clouds of ModelNet40 \cite{modelnet} and on
\sdf{} learned from meshes of Manifold40 \cite{subdivnet}.
We compare \algoname{} with other frameworks designed to process \nf{}s realized as Individual MLPs, such as DWSNet \cite{navon2023equivariant}.
These methods process the MLP weights of individual \nf{}s, implemented as SIREN networks \cite{siren}. The MLPs in our benchmark are initialized using the same random seed (see \cref{sec:same_init} for more details). We also tried to run other recent \textit{Individual} approaches, such as NFN \cite{zhou2023permutation} and NFT \cite{zhou2023neural}, but the training did not converge, probably because the SIREN MLPs used in our experiments are much larger than those used in their paper.
Moreover, we compare with \textit{Shared} frameworks where neural fields are realized by a shared network and a small latent vector or modulation, \ie{} DeepSDF \cite{deepsdf} and Functa \cite{functa}.
Whenever possible, we use the official code released by the authors to run the experiments.

As we can see from results reported in \cref{tab:neural_functional_classification}, Functa and \algoname{} achieve the best results with a large margin over other competitors, with the former slightly more accurate on \udf{}s while the latter on \sdf{}s.
However, as explained in the previous section, since our method does not rely on shared networks, it does not sacrifice the representation quality of \nf{}s and it is more suitable for real-world applications.

\begin{figure*}[t]
    \centering
    \includegraphics[width=0.8\linewidth]{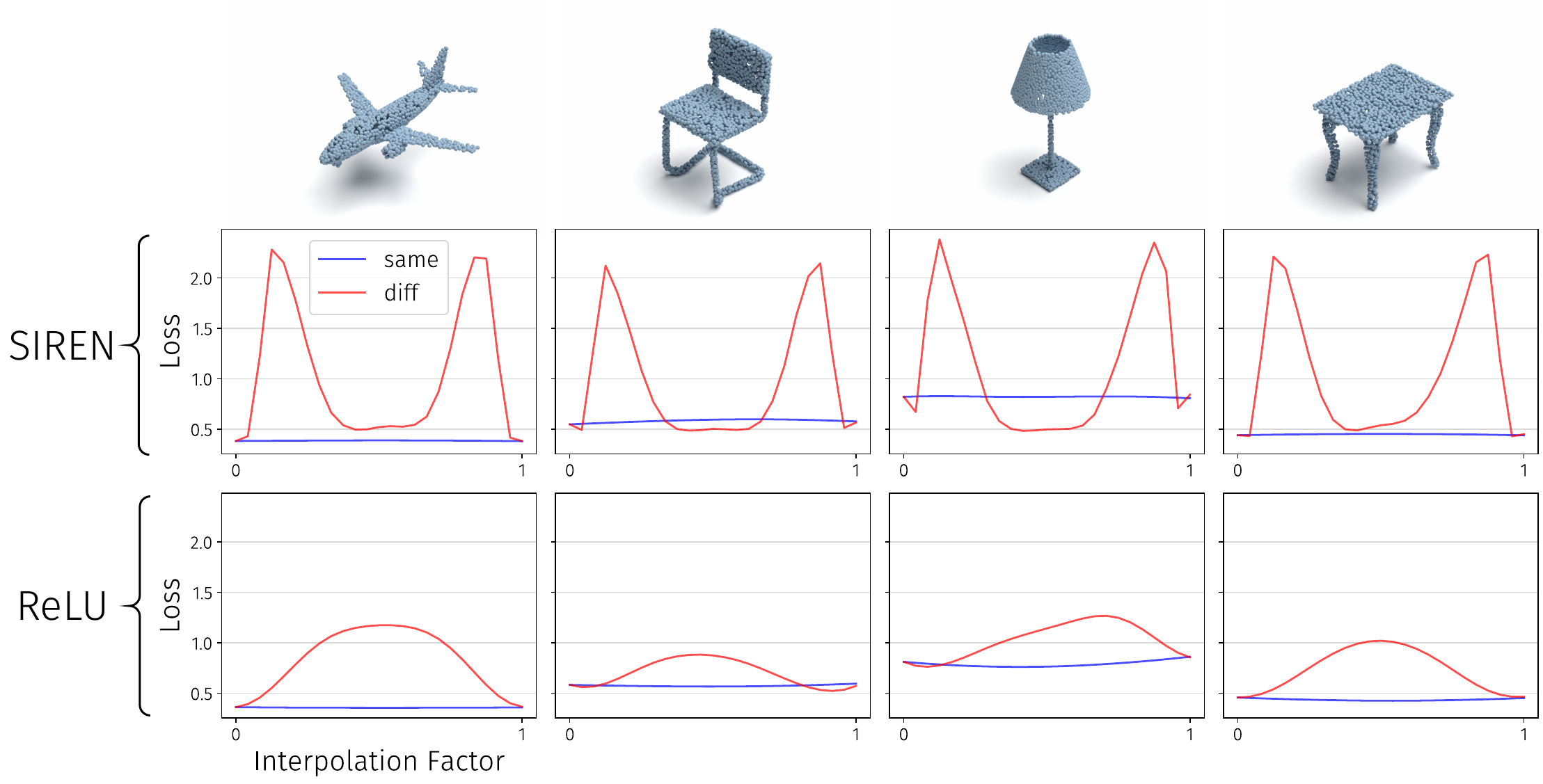}
    \caption{\textbf{Linear mode connectivity study.} Each plot shows the variation of the loss function over the same batch of points when interpolating between two \nf{}s representing the same shape. The red line describes the interpolation between \nf{}s initialized differently, whereas the blue line shows the interpolation between \nf{}s initialized with the same random vector.}
    \label{fig:lmc}
\end{figure*}

\section{Using the Same Initialization for \nf{}s}
\label{sec:same_init}

The need to align the multitude of \nf{}s that can approximate a given shape is a challenging research problem that has to be dealt with when using \nf{}s as input data. We empirically found that fixing the weights initialization to a shared random vector across \nf{}s is a viable and simple solution to this problem.

We report here an experiment to assess if the order of data or other sources of randomness arising while fitting \nf{}s do affect the repeatability of the embeddings computed by \algoname{}. We fitted 10 \nf{}s on the same discrete shape for 4 different chairs, \ie{} 40 \nf{}s in total. Then, we embed all of them with the pretrained \algoname{} encoder and compute the L2 distance between all pairs of embeddings. The block structure of the resulting distance matrix (see \cref{fig:init_distances}) highlights how, under the assumption of shared initialization, 
\algoname{} is repeatable across multiple fittings.

Seeking for a proof with a stronger theoretical foundation, we turn our attention to the recent work
\textit{git re-basin} \cite{rebasin}, where authors show that the loss landscape of neural networks contains (nearly) a single basin after accounting for all possible permutation symmetries of hidden units.
The intuition behind this finding is that, given two neural networks that were trained
with equivalent architectures but different random initializations, data orders, and potentially different hyperparameters or datasets, it is possible to find a permutation of the network's weights such that when linearly interpolating between their weights, all intermediate models enjoy performance similar to them -- a phenomenon denoted as \textit{linear mode connectivity}.

Intrigued by this finding, we conducted a study to assess whether initializing \nf{}s with the same random vector, which we found to be key to \algoname{} convergence, also leads to linear mode connectivity.
Thus, given one shape, we fitted it with two different \nf{}s, and then we interpolated linearly their weights, observing at each interpolation step the loss value obtained by the \textit{interpolated} \nf{} on the same batch of points.
We repeated the experiment twice for each shape, once initializing the \nf{}s with different random vectors and once with the same random vector.

The results of this experiment are reported for four different shapes in \cref{fig:lmc}. It is possible to note that, as shown by the blue curves, when interpolating between \nf{}s obtained from the same weights initialization, the loss value at each interpolation step is nearly identical to those of the boundary \nf{}s. On the contrary, the red curves highlight how there is no linear mode connectivity at all between \nf{}s obtained from different weights initializations.

\cite{rebasin} also proposes different algorithms to estimate the permutation needed to obtain linear mode connectivity between two networks. We applied the algorithm proposed in their paper in Section 3.2 (\textit{Matching Weights}) to our \nf{}s and observed the resulting permutations.
Remarkably, when applied to \nf{}s obtained from the same weights initialization, the retrieved permutations are identity matrices, both when the target \nf{}s represent the same shape and when they represent different ones.
Instead, the permutations obtained for \nf{}s obtained from different initializations are far from being identity matrices.

All these results favor the hypothesis that our technique of initializing \nf{}s with the same random vector leads to linear mode connectivity between different \nf{}s. We believe that the possibility of performing meaningful linear interpolation between the weights occupying the same positions across different \nf{}s can be interpreted by considering corresponding weights as carrying out the same role in terms of feature detection units, explaining why the \algoname{} encoder succeeds in processing the weights of our \nf{}s.



\begin{table}
    \centering
    \begin{tabular}{lc}
        \toprule
        Method & Test accuracy (\%)  \\
        \midrule
        \algoname{} SIREN & 87.0 \\
        \algoname{} ReLU & \textbf{88.1} \\
        \bottomrule
    \end{tabular}
    \caption{\textbf{Point cloud classification from \udf{} with \algoname{}.} Comparison of \algoname{} trained on SIRENs \cite{siren} vs ReLU MLPs on ModelNet40.}
    \label{tab:pcd-class-sine-relu}
\end{table}

The experiments in this section were conducted on \nf{} with sine and ReLU activation functions, as those are the activations used throughout this paper. To further validate the applicability of our method to SIRENs and ReLU \nf{}s, we show in \cref{tab:pcd-class-sine-relu} the comparable results obtained by classifying \algoname{} embeddings of \nf{}s with different activation functions.

\section{Limitations}
\label{sec:limitations}

We point out three main limitations of our approach:
\begin{enumerate*}[label=(\roman*)]
  \item although \nf{}s capture continuous geometric cues, in some cases deep learning on \algoname{} embeddings achieve results inferior to state-of-the-art solutions that work on specific discrete representations;
  \item there is no obvious way to perform online data augmentation on shapes represented as \nf{}s by directly altering their weights;
  \item when processing NeRFs, appearance and geometry are intrinsically entangled in our framework, both in NeRFs as an architecture, where a single MLP predicts both color and density, and in the method we use to process them, which does not take any measure aimed at disentangling geometric and texture information. As a result, \algoname{} could not be applied in scenarios where one needs to control how much appearance and geometry independently affect the outcome.
\end{enumerate*}
We plan to investigate these shortcomings in future works.

\section{Concluding Remarks}
\label{sec:conclusion}

We have shown that it is possible to apply deep learning on individual \nf{}s representing 3D shapes and object-centric radiance fields.
Our approach leverages a task-agnostic encoder which embeds \nf{}s into compact and meaningful latent codes without accessing the underlying function.
%
We have shown that these embeddings can be fed to standard deep-learning machinery to solve various tasks effectively.
Moreover, we have introduced the first benchmark for the task of \nf{} classification, showing that our proposal obtains the best score (on par with Functa \cite{functa}) while preserving the ability to reconstruct the input dataset with high quality.

In the future, we plan to go beyond \nf{}s of 3D objects by applying \algoname{} to \nf{}s encoding 3D scenes and other input modalities, like images or audio.
We will also investigate weight-space symmetries \cite{entezari2021role} as a different path to favor the alignment of weights across \nf{}s, possibly while exploring the effect of activation functions other than sine and ReLU.

We reckon that our work may foster the adoption of \nf{}s as a unified 3D representation, overcoming the current fragmentation of 3D structures and processing architectures.

\bibliographystyle{IEEEtran}
\bibliography{references}

\begin{IEEEbiography}
[{\includegraphics[width=1in,height=1.25in,clip,keepaspectratio]{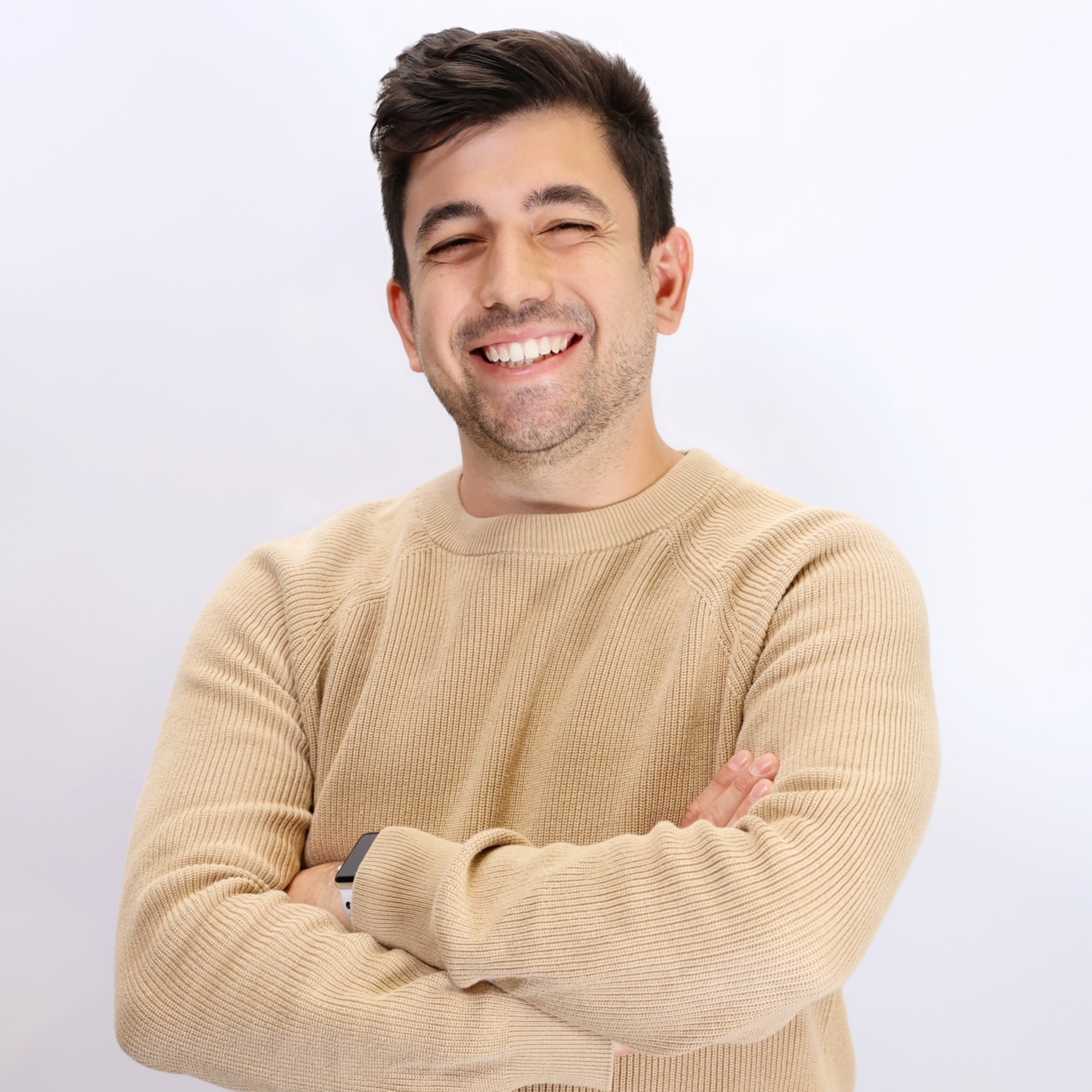}}]
{Pierluigi Zama Ramirez} received his PhD in Computer Science and Engineering in 2021. He has been a Research Intern at Google for 6 months and is currently a Post-Doc at the University of Bologna. He co-authored more than 20 publications on computer vision research topics such as semantic segmentation, depth estimation, optical flow, domain adaptation, virtual reality, and 3D computer vision.
\end{IEEEbiography}
\begin{IEEEbiography}[{\includegraphics[width=1in,height=1.25in,clip,keepaspectratio]{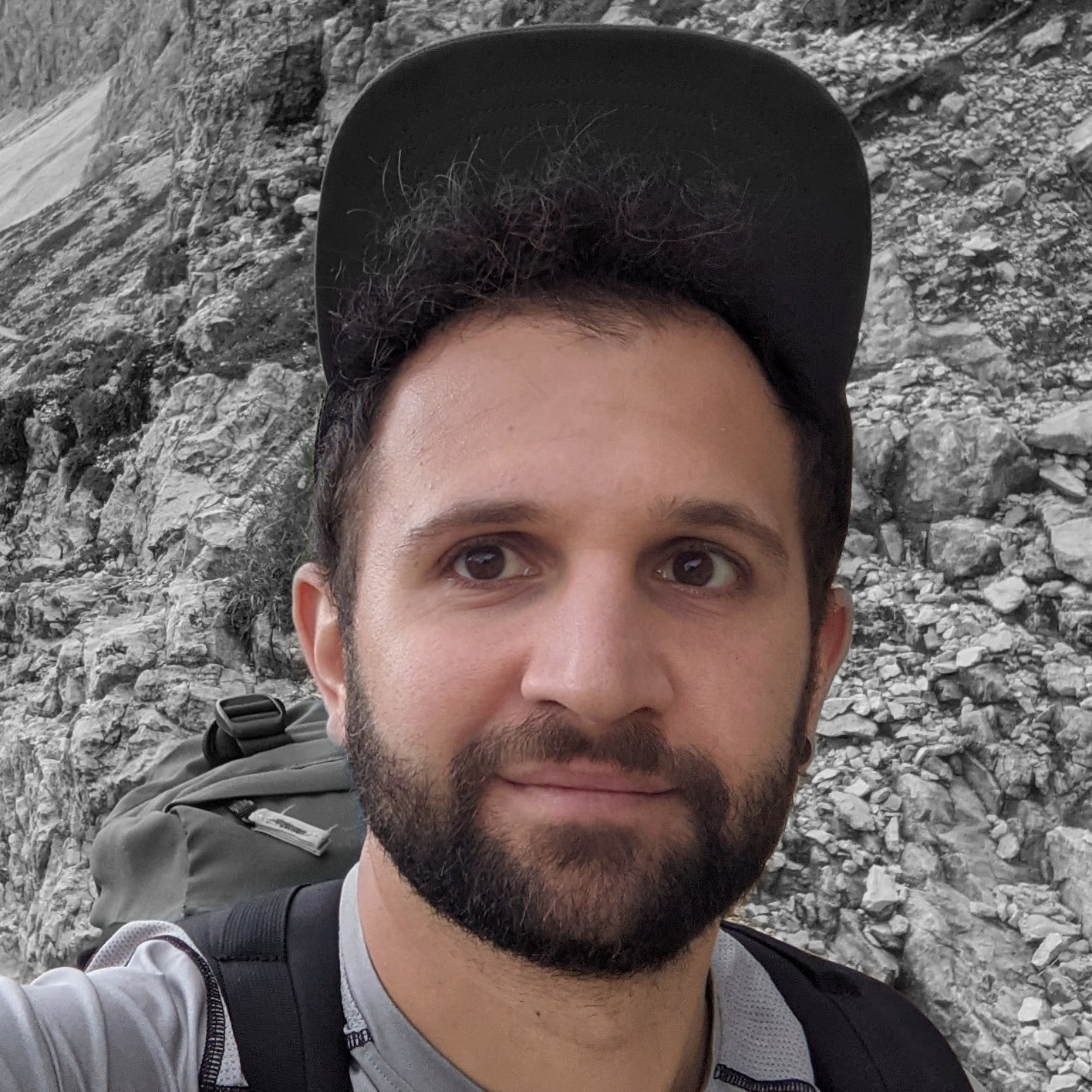}}]{Luca De Luigi} received his PhD in Computer Science and Engineering in 2023. He is currently a computer vision engineer at eyecan.ai. His research focuses on deep learning for computer vision problems, especially in 3D geometry and neural fields.
\end{IEEEbiography}
\begin{IEEEbiography}
[{\includegraphics[width=1in,height=1.25in,clip,keepaspectratio]{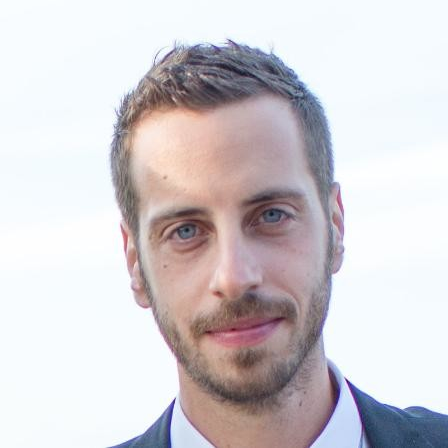}}]
{Daniele Sirocchi}
received his master’s degree in Artificial Intelligence in 2023. He currently works as a software engineer, mainly focused on creating applications that blend various technologies, encompassing a wide spectrum of technological domains. His research interests primarily revolve around machine and deep learning for computer vision tasks, where he extensively engages with neural fields.
\end{IEEEbiography}
\begin{IEEEbiography}
[{\includegraphics[width=1in,height=1.25in,clip,keepaspectratio]{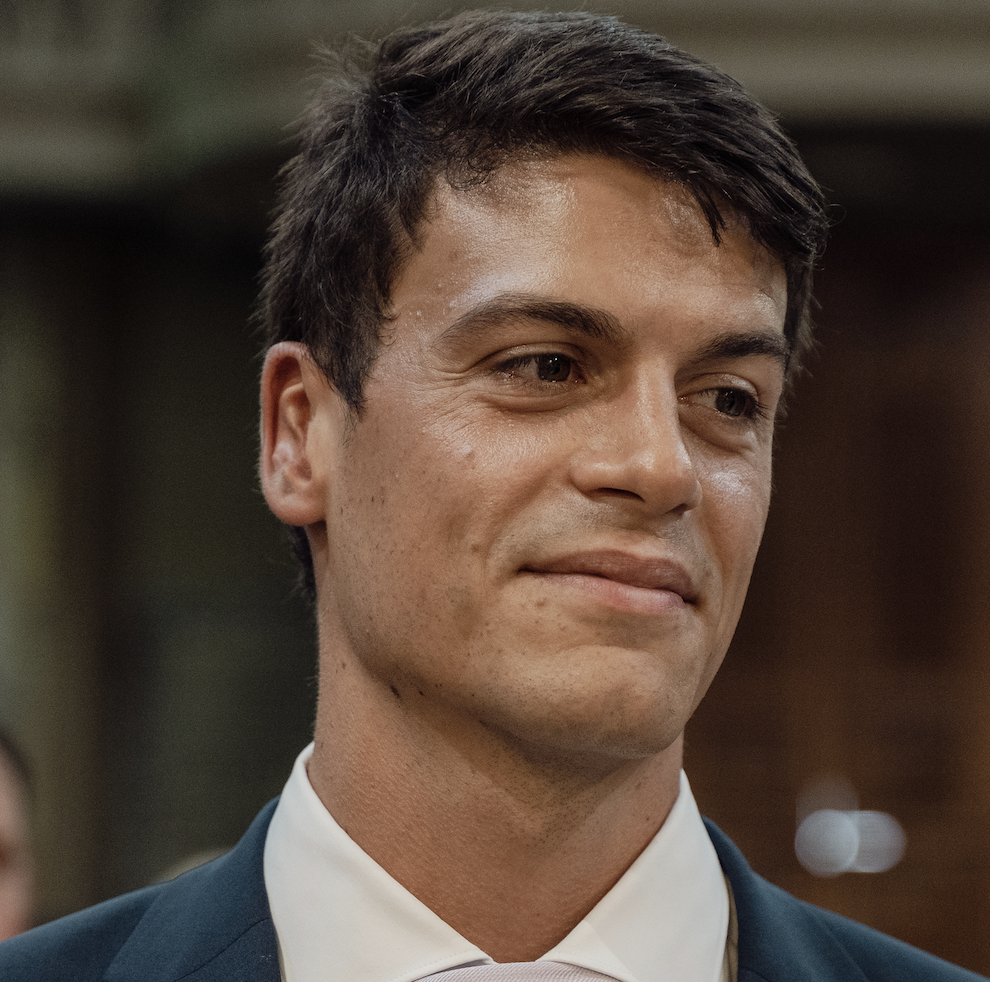}}]
{Adriano Cardace} is a third-year PhD student at the Computer Vision Laboratory (CVLab), University of Bologna. He has been a Research Intern at Mitsubishi Electric Research Laboratories (MERL) and authored several research papers covering a range of subjects, including semantic segmentation, Domain Adaptation, and Neural Fields.
\end{IEEEbiography}
\begin{IEEEbiography}[{\includegraphics[width=1in,height=1.25in,clip,keepaspectratio]{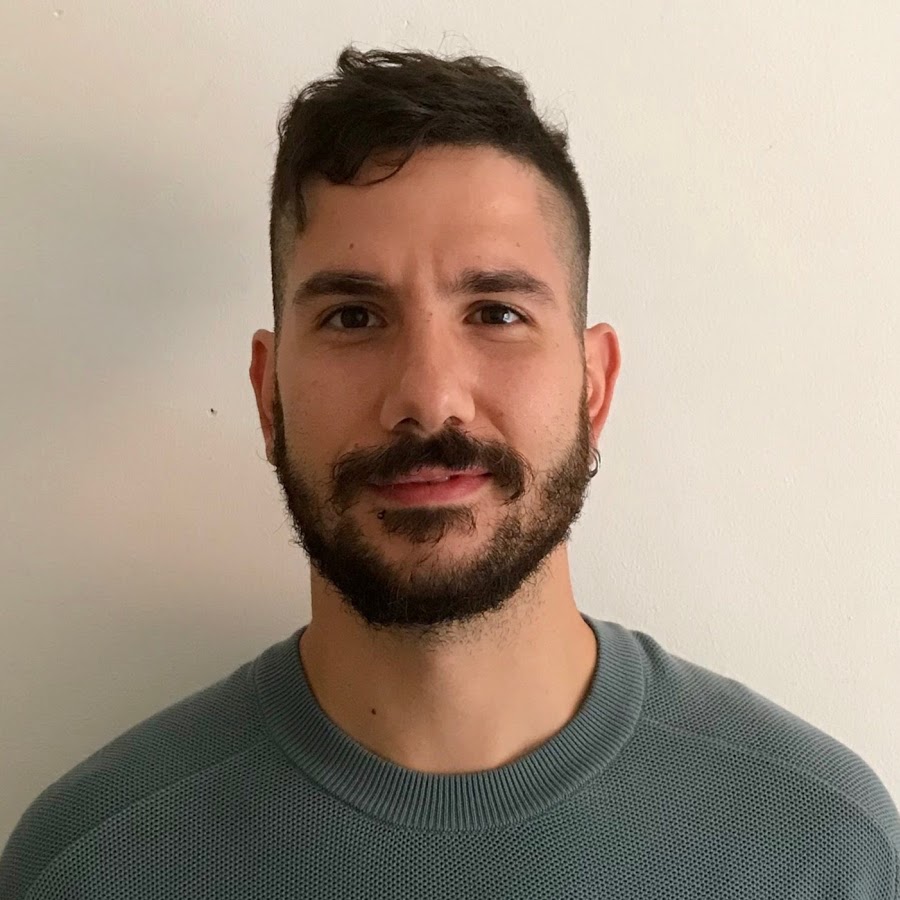}}]{Riccardo Spezialetti}
received his PhD degree in Computer Science and Engineering from University of Bologna in 2020. After two years as a Post-doc researcher at the Department of Computer Science and Engineering, University of Bologna, he recently joined eyecan.ai as a computer vision engineer. His research interest concerns machine/deep learning for 3D computer vision problems.
\end{IEEEbiography}
\begin{IEEEbiography}
[{\includegraphics[width=1in,height=1.25in,clip,keepaspectratio]{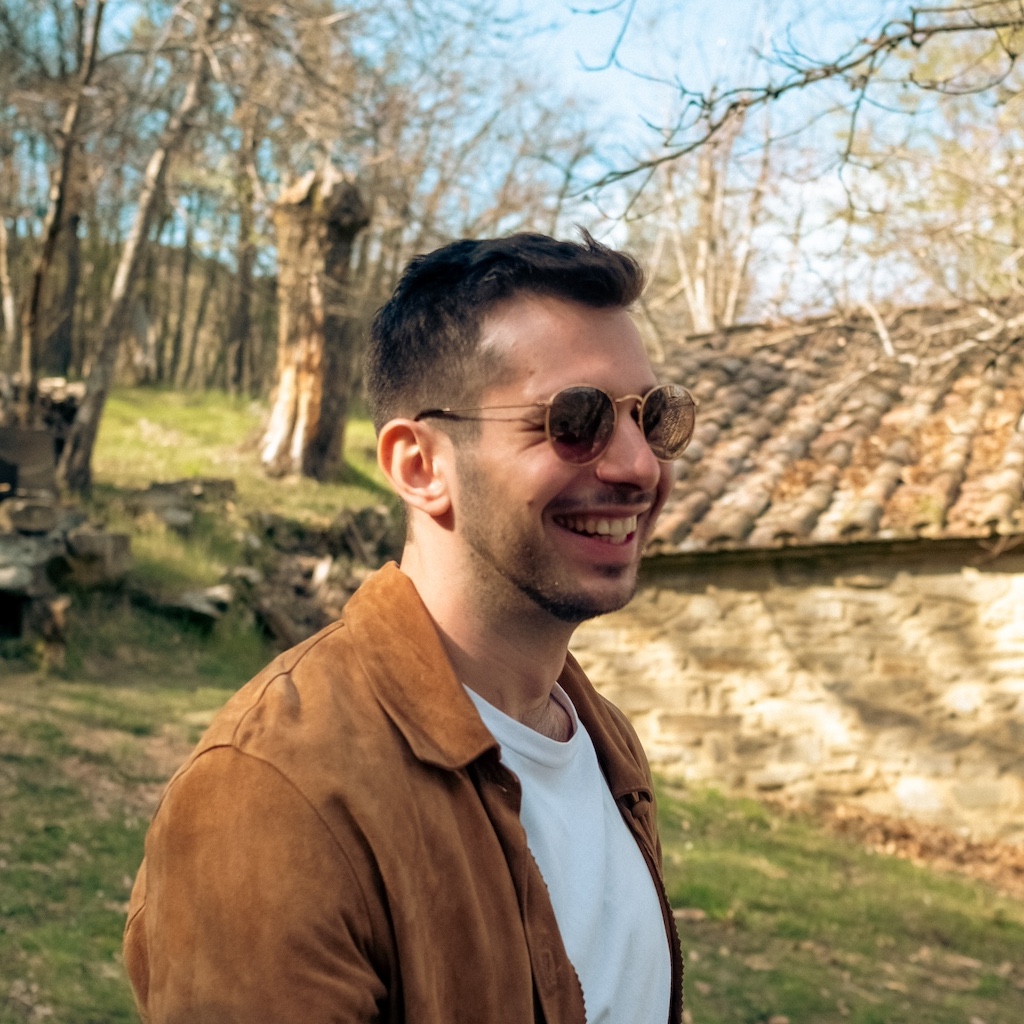}}]
{Francesco Ballerini} received his Master's degree in Artificial Intelligence in 2023 and is currently a PhD student in Computer Science and Engineering, both at the University of Bologna. His main research interests include the study of neural architectures aimed at processing neural fields, specifically those representing 3D data, with a focus on permutation symmetries and the effect of initialization on the input neural fields themselves.
\end{IEEEbiography}
\begin{IEEEbiography}
[{\includegraphics[width=1in,height=1.25in,clip,keepaspectratio]{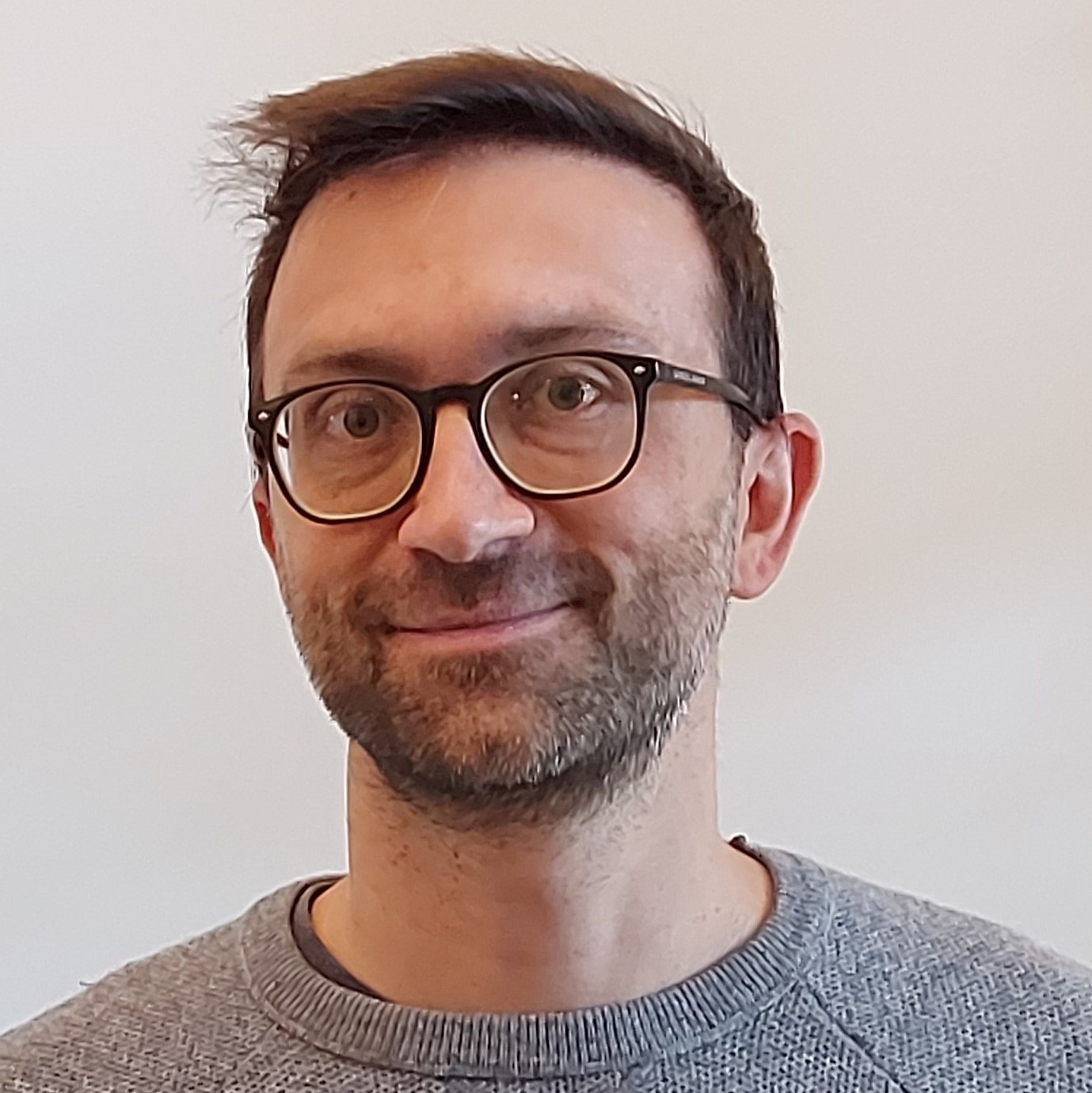}}]{Samuele Salti} is currently an associate professor at the Department of Computer Science and Engineering (DISI) of the University of Bologna, Italy.  His main research interest is computer vision, mainly 3D computer vision and machine/deep learning applied to computer vision problems.
Dr. Salti has co-authored more than 60 publications and 8 international patents. In 2020, he co-founded the start-up eyecan.ai. 
\end{IEEEbiography}
\begin{IEEEbiography}
[{\includegraphics[width=1in,height=1.25in,clip,keepaspectratio]{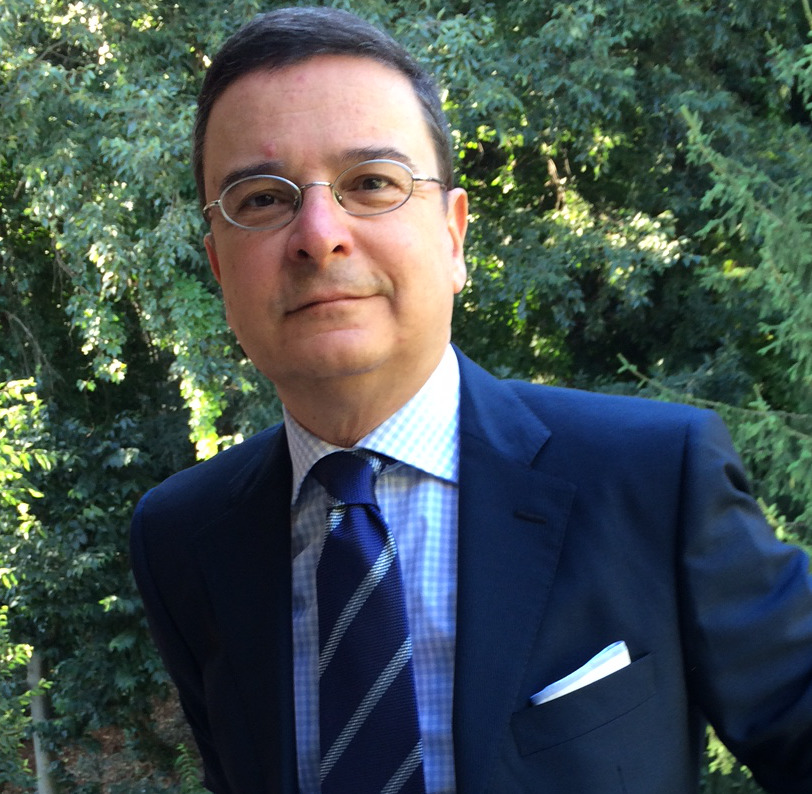}}]
{Luigi Di Stefano}
received a PhD degree in electronic engineering and computer science from the University of Bologna in 1994. He is a full professor at the Department of Computer Science and Engineering, University of Bologna, where he founded and led the Computer Vision Laboratory (CVLab). His research interests include image processing, computer vision, and machine/deep learning. He is the author of more than 150 papers and several patents. He has been a scientific consultant for major computer vision and machine learning companies.  He is a member of the IEEE Computer Society and the IAPR-IC.
\end{IEEEbiography}

\end{document}